\documentclass{article} 
\usepackage{iclr2027_conference,times}

\newcommand{\R}{\mathbb{R}}
\newcommand{\E}{\mathbb{E}}

\newcommand{\cL}{\mathcal{L}}

\newcommand{\din}{d_{\mathrm{in}}}
\newcommand{\dout}{d_{\mathrm{out}}}

\usepackage{amsmath,amsfonts,bm}

\def\eqref#1{equation~\ref{#1}}

\def\1{\bm{1}}

\DeclareMathAlphabet{\mathsfit}{\encodingdefault}{\sfdefault}{m}{sl}
\SetMathAlphabet{\mathsfit}{bold}{\encodingdefault}{\sfdefault}{bx}{n}

\usepackage{enumitem}
\usepackage{hyperref}
\usepackage{booktabs}
\usepackage{url}
\usepackage{graphicx}
\usepackage{multirow}
\usepackage{comment}

\title{Weights Read and Write Features: Scalable Parameter Decomposition Grounded in Activation Space}

\author{Tue M. Cao$^{1}$, Lisiane Pruinelli$^{2}$, My T. Thai$^{1}$\thanks{Corresponding author.} \\
$^{1}$Department of Computer and Information Science and Engineering, University of Florida \\
$^{2}$College of Nursing, University of Florida \\
Gainesville, FL 32611, USA \\
\texttt{\{caotue, lisianepruinelli\}@ufl.edu}, \texttt{mythai@cise.ufl.edu}
}

\iclrfinalcopy 
\begin{document}

\maketitle
\lhead{} 

\begin{abstract}
Activation space and parameter space provide complementary views of model
computation. Activations represent information, while weights read, transform,
and write that information. Yet existing interpretability methods largely study the two spaces
separately, leaving the connection between represented information and
parameter-level computation underexplored. We introduce Activation-Supported
Parameter Decomposition (ASPD), which jointly decomposes activation and
parameter spaces and grounds each learned weight component in the activation
features it reads or writes. This grounding constrains otherwise non-unique
parameter decompositions using the model's internal activations, while an
internal reconstruction objective provides a local learning signal at the
weight matrix being analyzed. Together, these properties enable scalable,
interpretable, and causally editable parameter decomposition in pretrained
large language models, demonstrated on Qwen-3-8B. The learned read--write
components can also be composed into parameter-level mechanism circuits. We
use ASPD to recover mechanisms underlying the classic IOI circuit and trace
semantic transformations through model weights. \footnote{Code available at \url{https://github.com/tue147/weights-read-write}.}
\end{abstract}

\section{Introduction}
\label{sec:intro}
Understanding model computation requires reasoning about two complementary
objects, the information represented in activation space
 and the mechanisms implemented in
parameter space. Activations represent
information, while weight matrices read that information, transform it, and
write new information into downstream activations.  Yet most interpretability methods analyze these
spaces separately \citep{on_biology, vpd, open_problem}.


Activation-space decomposition, particularly Sparse Autoencoders
\citep{monosemanticity, gemma_scope, lieberum2024gemma}, has revealed sparse
and interpretable features and enabled applications such as feature circuits
\citep{feature_circuit, transcoder_circuit, on_biology} and model diffing
\citep{modeldiff_application}. However, identifying an activation feature does
not directly reveal which weights create, consume, or transform it. Parameter
decomposition methods \citep{apd, spd, l3d, tpd, vpd} address the
complementary problem by decomposing weight matrices into localized components,
but the learned weight components are not explicitly grounded in the
activation features on which the model operates. Existing methods for
unsupervised parameter-mechanism discovery also remain difficult to scale to
large pretrained models \citep{vpd}.

We argue that these limitations are closely related. A weight matrix
generally admits many possible decompositions, while the weights alone do not
specify which components correspond to mechanisms actually exercised by the
model on its activation distribution. Internal activations provide this
missing constraint by revealing what information is present when a weight
component is used and what information its computation produces. They also
provide a local learning signal for an internal weight matrix, avoiding the
need to infer its mechanisms only through changes at the final model output \citep{vpd}.
Activation grounding therefore serves not only to interpret learned weight
components, but also to constrain the parameter decomposition and make it
practical in deeper models.

Based on this view, we introduce Activation-Supported Parameter Decomposition
(ASPD), which jointly decomposes activation and parameter spaces. We formulate
activation-grounded parameter decomposition in terms of sparse causal
relationships between weight components and activation features, so that a
component can be characterized by the information it reads or writes. Directly
optimizing all feature--component interventions, however, is prohibitively
expensive. ASPD provides a tractable surrogate by using a shared sparse
representation that jointly defines activation features and gates the
corresponding weight components. An internal reconstruction objective further
requires these components to reproduce the transformation of the target weight
matrix directly at its output. Together, activation grounding and local
reconstruction yield interpretable and causally editable weight components in
multi-billion-parameter language models, including Qwen-3-8B. 

\begin{figure}[t]
    \centering
    \vspace{-5mm}
    \includegraphics[width=\linewidth]{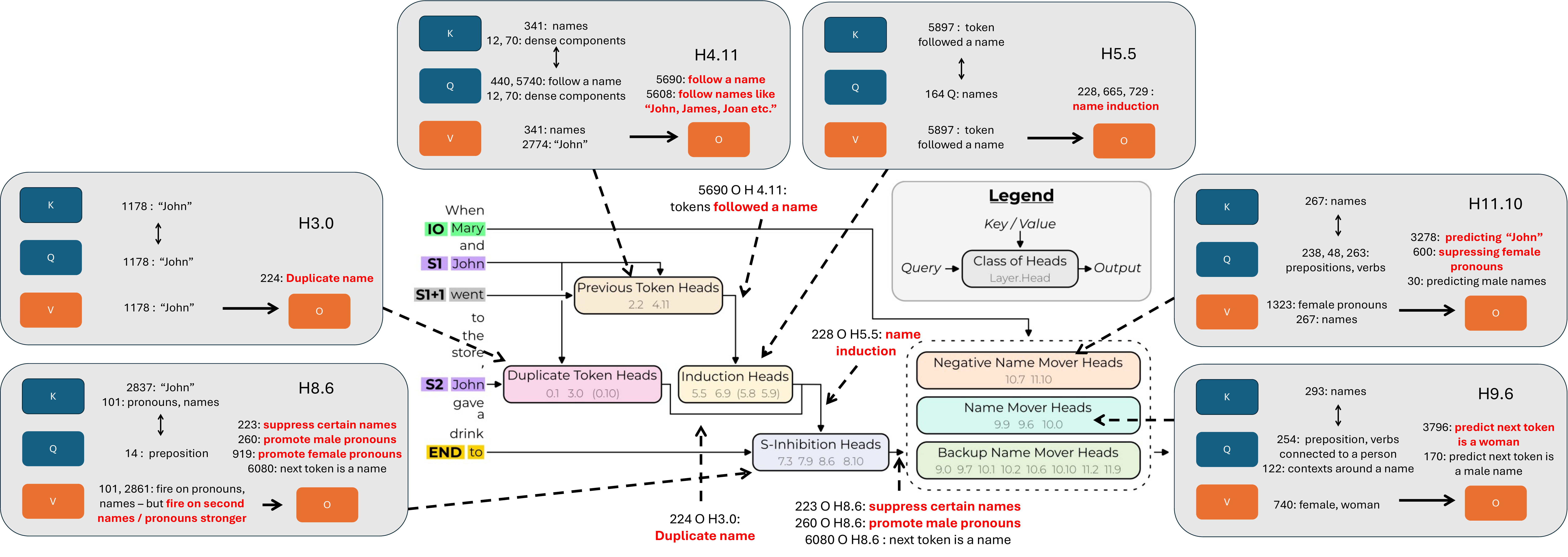}
    \caption{
    Parameter-level mechanisms recovered for the IOI circuit \citep{ioi}.
    ASPD identifies weight components implementing known head-level
    computations and connects them through read--write interactions across
    heads. For example, the duplicate-name component $224O$ of H3.0 and the
    induction component $228O$ of H5.5 provide inputs to the downstream
    S-inhibition mechanism in H8.6.
    }
    \label{fig:ioi_circuit_mech}
    \vspace{-4mm}
\end{figure}

The read--write formulation also allows individual weight components to be
composed into larger parameter-level mechanisms. A component that writes
information along a particular direction can influence downstream components
that read that information. We use these interactions to construct
parameter-level mechanism circuits and apply ASPD to the well-studied Indirect
Object Identification (IOI) circuit \citep{ioi}. Without training on circuit labels,
ASPD recovers weight components corresponding to known computations including
induction, duplicate-token detection, S-inhibition, and name movement, and
reveals how these mechanisms communicate across heads. Figure~\ref{fig:ioi_circuit_mech} previews the resulting parameter-level IOI
circuit; we analyze these mechanisms in detail in
Section~\ref{sec:ioi_circuit}. 
We also use ASPD to
trace how semantic information is read, transformed, and written through model
weights.

Our contributions are:
\begin{itemize}[noitemsep,topsep=0pt,parsep=0pt,partopsep=0pt,leftmargin=*]
    \item We introduce activation-grounded parameter decomposition, which
    characterizes weight components through the activation features they read
    and write, and propose ASPD to jointly learn activation features and weight
    mechanisms.

    \item ASPD replaces expensive pairwise feature--component interventions
    with a shared sparse representation and an internal reconstruction signal,
    enabling interpretable and causally editable parameter decomposition in
    models up to Qwen-3-8B.

    \item We develop a read--write interaction framework for composing weight
components into parameter-level mechanism circuits, and demonstrate it on the
IOI circuit and semantic transformations through model weights.
\end{itemize}

\section{Related Work}
\label{sec:related_works}

\textbf{Activation-space interpretability.}
A large body of work studies model representations through hidden activations.
SAEs \citep{monosemanticity, gemma_scope, lieberum2024gemma} and Transcoders
\citep{transcoder_circuit} decompose activations into sparse, interpretable
features, enabling applications such as feature circuits
\citep{feature_circuit}, model diffing \citep{diffing, model_diffing}, and
analysis of activation-space geometry \citep{geometry, saegeometry, tree_sae}.
Other approaches inspect or explain hidden representations through vocabulary
projection or learned activation explanations
\citep{logitlens, tuned_lens, patchscope, activation_oracle}.
These methods primarily characterize what information is represented in
activation space. Our goal is instead to connect such representations to the
parameter components that read, transform, and write them, and thus better explaining the underlying computation.

\textbf{Parameter-space interpretability.}
Several lines of work manipulate or interpret model parameters. Knowledge
editing \citep{weight_edit, think_weight_edit, principled_weight_edit} and
weight steering \citep{weight_steer} modify specific knowledge or behaviors,
while vocabulary-based analyses
\citep{kv_mlp, feedforward_vocab} associate individual parameter
directions with output tokens. Other approaches construct models with
interpretable weight structure by design
\citep{bilinear_mlp, weight_sparse_trans}. More directly related to our work,
parameter decomposition seeks to recover computational structure from existing
weights. Early studies use SVD to identify interpretable directions
\citep{svd_weight_1, svd_weight_2} or sparse weight decomposition
\citep{swd} targets sparse parameter representations for circuit extraction but these methods are limited in interpretability.  Recent methods \citep{apd, spd, l3d, tpd, vpd} learn sparse components
intended to capture interpretable mechanisms. However, these components are learned
primarily within parameter space and are not explicitly grounded in the
activation-space features on which the model operates. Moreover, methods aimed
at unsupervised recovery of interpretable parameter mechanisms remain difficult
to scale to large models \citep{vpd}. ASPD addresses both
issues by using activation-space structure as semantic grounding and as an
internal learning signal for parameter decomposition.

\textbf{Connecting activation and parameter spaces.} The idea of explaining both spaces together is under-explored. \citet{precise_in_param} use directions from pretrained SAEs \citep{gemma_scope} to guide parameter-space editing, while \citet{lowrank_finetune_for_sae} modify model parameters to improve SAE representations. They demonstrate useful interactions between
activations and weights, but cannot provide explanation for the underlying mechanisms of a model. 
In contrast, ASPD learns parameter components according to the activation-space features they read or write, providing a direct interpretability method to explain the computation mechanisms.

\section{Activation-Grounded Parameter Decomposition}
\label{sec:method}

\subsection{Weight Components as Read--Write Mechanisms}
\label{sec:read_write}

Let $W\in\mathbb{R}^{d_{\mathrm{out}}\times d_{\mathrm{in}}}$ be a weight
matrix and $X=(x_1,\ldots,x_T)$ its input activations for a length-$T$ token
sequence, where $x_t\in\mathbb{R}^{d_{\mathrm{in}}}$. Its output is
$Y=(y_1,\ldots,y_T)$ with
$y_t=Wx_t\in\mathbb{R}^{d_{\mathrm{out}}}$.
Parameter decomposition represents the computation of $W$ using $C$ rank-1
weight components
\[
P_c=u_cv_c^\top,\qquad
u_c\in\mathbb{R}^{d_{\mathrm{out}}},
\quad
v_c\in\mathbb{R}^{d_{\mathrm{in}}},
\quad c=1,\ldots,C,
\]
together with a causal importance function $g_{t,c}(X)$ indicating when component $c$
participates in the computation \citep{spd,vpd}; $g_{t,c}(X)>0$ means the component is important for the computation at token $t$, and $g_{t,c}(X)=0$ means otherwise. We say component $c$ \textit{fires} iff $g_{t,c}>0$. We define
\begin{equation}
    e_{t,c}=g_{t,c}(X)(v_c^\top x_t),
    \qquad
    \hat y_t=\sum_{c=1}^{C} e_{t,c}u_c ,
    \label{eq:component_computation}
\end{equation}
where $e_{t,c}$ is the contribution of component $c$ at position $t$
and $\hat y_t$ is the reconstructed output of $W$.

This gives each weight component a read--write interpretation. The direction
$v_c$ determines what information the component reads, $g_{t,c}$ determines
when it is active, and $u_c$ determines what information it writes to the
downstream activation. We therefore view a weight component as a
context-dependent mechanism characterized by $(v_c,g_c,u_c)$ rather than only
as a rank-1 matrix.

\subsection{Activation-Grounded Weight Components}
\label{sec:act_decomp_param_decomp}

A weight matrix can admit many possible decompositions, while the weights alone
do not specify which components correspond to mechanisms exercised by the
model. We use activation space to constrain this ambiguity. Let
$R=R(X)=(r_1,\ldots,r_T)$ 
denote the input activations to an activation-space decomposition, where
$r_t\in\mathbb{R}^{d_{\mathrm{act}}}$. Let
$a(r_t)=(a_1(r_t),\ldots,a_F(r_t))\in\mathbb{R}^{F}$ denote a decomposition
of this activation space into $F$ features, such as an SAE \citep{monosemanticity}.

We define the relationship between weight components and activation features
causally. When $R$ is downstream of $W$, the write effect of component $c$ on
feature $i$ is
\begin{equation}
    A^{\mathrm{write}}_{i,c}(X,t)
    =
    a_i(r_t)
    -
    a_i(r_t\mid g_{t,c} \leftarrow 0),
    \label{eq:write_effect}
\end{equation}
where $a_i(r_t\mid g_{t,c}\leftarrow 0)$ denotes the feature activation obtained
when the contribution of weight component $c$ is set to zero while all other
components are left unchanged. Eq. (\ref{eq:write_effect}) measures how much feature $i$ changes when component $c$ is removed.
When $R$ is upstream of $W$, the read effect is
\begin{equation}
    A^{\mathrm{read}}_{i,c}(X,t)
    =
    g_{t,c}(X)
    -
    g_{t,c}\!\left(X\mid a_i(r_t)\leftarrow 0\right),
    \label{eq:read_effect}
\end{equation}
which measures how much the use of component $c$ changes when feature $i$ is
removed.

An activation-grounded parameter decomposition should concentrate these causal
relationships on a small set of activation features. The ideal grounding objective is
\begin{equation}
    \mathcal{L}_{\mathrm{ground}}
    =
    \mathcal{L}_c
    +
    \frac{\lambda_{\mathrm{ground}}}{CF}
    \mathbb{E}_{X}
    \left[
    \frac{1}{T}
    \sum_t\sum_i\sum_c
    |A_{i,c}(X,t)|
    \right],
    \label{eq:causal_ground}
\end{equation}
where $\mathcal{L}_c$ denotes some constraining loss, $\mathbb{E}_X$ denotes expectation over the training data distribution, $A_{i,c}$ denotes the appropriate read or write effect and
$\lambda_{\mathrm{ground}}$ controls the grounding objective. Sparse causal
relationships associate each weight component with localized activation-space
mechanisms and characterize it through the features it reads or writes.

Directly optimizing Equation~\ref{eq:causal_ground} would require
interventions over all $F\times C$ feature--component pairs at every training
step and is computationally prohibitive. Our ASPD therefore replaces this explicit
causal objective with a structural surrogate. We align each weight component
with one activation feature by using the same sparse latent coordinate to
represent feature $c$ and control the gate of component $c$.
The shared representation forces each component to either read from or write to a feature in the activation space, making each component has a localized and interpretable mechanism.

\subsection{ASPD for Joint Activation--Parameter Decomposition}
\label{sec:aspd}

Activation-Supported Parameter Decomposition (ASPD) realizes this structural
surrogate by jointly learning activation features and weight components. We
set $C=F$ and learn a shared sparse encoder
$
g^s:\mathbb{R}^{T\times d_{\mathrm{act}}}
\rightarrow\mathbb{R}^{T\times C}
$
and an activation decoder
$
d:\mathbb{R}^{T\times C}
\rightarrow\mathbb{R}^{T\times d_{\mathrm{act}}}.
$
The same latent coordinate defines activation feature $c$ and gates weight
component $c$:
\begin{equation}
    a(R) := g^s(R),
    \qquad
    g_{t,c}(X)
    :=
    \phi\!\left(g^s_{t,c}(R)\right),
    \label{eq:shared_gate}
\end{equation}
where $g^s_{t,c}(R)$ is the activation of feature $c$ at position $t$ and
$\phi:\mathbb{R}\rightarrow\mathbb{R}$ maps a feature activation to its
component gate. Thus, the shared representation provides the structural
surrogate for the read/write grounding defined in
Equations~\ref{eq:write_effect}--\ref{eq:read_effect}, while
$u_cv_c^\top$ specifies the weight-space transformation performed by the
corresponding component.

ASPD jointly optimizes
\begin{align}
    \mathcal{L}_{\mathrm{ASPD}}
    &=
    \mathcal{L}_{\mathrm{internal}}
    +
    \lambda_{\mathrm{act}}\mathcal{L}_{\mathrm{act}}
    +
    \lambda_{\mathrm{sparse}}\mathcal{L}_{\mathrm{sparse}}(g^s),
    \label{eq:jointly}\\
    \mathcal{L}_{\mathrm{internal}}
    &=
    \mathbb{E}_{X}
    \left[
    \frac{1}{T}\sum_{t=1}^{T}
    \left\|y_t-\hat y_t\right\|_2^2
    \right],
    \label{eq:internal}\\
    \mathcal{L}_{\mathrm{act}}
    &=
    \mathbb{E}_{X}
    \left[
    \frac{1}{T}
    \left\|R-d(g^s(R))\right\|_F^2
    \right],
    \label{eq:activation_reconstruction}
\end{align}
where 
$\mathcal{L}_{\mathrm{sparse}}$ encourages sparse activation of the shared
representation, and $\lambda_{\mathrm{act}}$ and
$\lambda_{\mathrm{sparse}}$ control the corresponding objectives.

The two reconstruction objectives serve complementary purposes.
$\mathcal{L}_{\mathrm{act}}$ forces the shared coordinates to form an
activation-space feature decomposition, grounding the weight components in
interpretable activation features. $\mathcal{L}_{\mathrm{internal}}$ requires
the gated components to reproduce the internal transformation performed by
$W$ on the model's activations. 
Because the gates depend on $R$, ASPD learns a sparse, data-conditioned decomposition of the computation induced by $W$, and each learned weight-space component $P_c=u_cv_c^\top$ that can be inspected or edited directly.

\textbf{Implication.} Jointly decomposing both activation and parameter space allows us to leverage many advancements in activation decomposition for parameter decomposition. One potential application is \textit{Parameter Diffing} (identifying what changed in the \textit{mechanisms} of the fine-tuned compared to the base model) by jointly decomposing activation diffing \citep{diffing, model_diffing, modeldiff_application} and parameter decomposition. We leave this direction for future work.

{\bf Relation to Transcoders.}
Under our formulation, Transcoder \citep{transcoder_circuit} can be recovered as a special case of
the formulation, and therefore, is a parameter decomposition method. We term
this interpretation PD Transcoder and give the full correspondence in
Appendix~\ref{sec:trans_reform}. ASPD generalizes this construction by decoupling the weight read
direction from the activation-feature encoder while preserving activation-space
semantics through $\mathcal{L}_{\mathrm{act}}$.

{\bf Internal learning signal.}
Prior parameter decomposition methods such as VPD were demonstrated on a
four-layer model and partly supervise internal components through constraints
on the final model output under component ablation \citep{vpd}.
Equation~\ref{eq:internal} instead supervises the decomposition directly at the
output of the weight matrix being analyzed, avoiding the need to propagate the
learning signal through all subsequent layers. This local supervision improves
decomposition quality in deeper and larger pretrained models. Consistently,
adding $\mathcal{L}_{\mathrm{internal}}$ to VPD improves its interpretability
and diversity, although it remains weaker than ASPD in Section~\ref{sec:exp}.

{\bf Practical instantiation.}
In our implementation, $g^s$ is a per-token BatchTopK encoder
\citep{batchtopk}. We use $\phi(s)=\mathbb{I}[s>0]$
where $\mathbb{I}[\cdot]$ denotes the indicator function, so each weight
component is either active or ablated at a token. BatchTopK enforces the target
sparsity directly; when used, the explicit
$\mathcal{L}_{\mathrm{sparse}}$ penalty is not required. We ground the shared
representation in residual-stream activations, we provide discussion for this choice in Appendix \ref{sec:discussion_gs}. Full architectural choices,
grounding locations, and training details are provided in Appendix~\ref{sec:training_details}.

{\bf VPD objectives.}
VPD additionally uses $\mathcal{L}_{\mathrm{param}}, \mathcal{L}_{\mathrm{ablate}}$ (full formulation in Appendix \ref{sec:background}) to encourage components that remain modular under independent ablations \citep{vpd}. Although we can trivially adapt these losses into our solution, however, in Appendix~\ref{sec:ablate_param}, we evaluate these objectives and finds mixed or negative effects, so
we omit them from ASPD.

\subsection{From Weight Components to Parameter-Level Mechanism Circuits}
\label{sec:mechanism_composition}

Model mechanisms generally involve interactions among weight components across
matrices and layers. The read--write formulation provides a natural way to
compose them. From Equation~\ref{eq:component_computation}, component $c_1$
writes $e_{t,c_1}u_{c_1}$ to the downstream activation, while a subsequent
component $c_2$ reads along $v_{c_2}$. For compatible sequential components,
including $OV$, $MLP_{in}, MLP_{out}$ matrices, and cross-layer residual interactions, we
score their interaction by
\begin{equation}
    \mathrm{Interact}(c_1,c_2)
    =
    \mathbb{E}_{X,t}
    \left[
    g_{t,c_2}(X)\,e_{t,c_1}
    \right]
    \left\langle u_{c_1},v_{c_2}\right\rangle ,
    \label{eq:component_interaction}
\end{equation}
where the expectation averages over examples and token positions in the
analysis corpus. The first factor measures whether the two components
participate on the same data, while
$\langle u_{c_1},v_{c_2}\rangle$ measures whether the information written by
$c_1$ aligns with the direction read by $c_2$. We use this score to rank
candidate computational dependencies between weight components.

For attention layers, $Q$, $K$, $V$, and $O$ denote the query, key, value, and
output weight matrices. $QK$ interactions require a head-specific composition
because both components write into query and key spaces rather than forming a
sequential write--read pair. The corresponding $QK$, $OV$, MLP, and
cross-layer interaction forms used in our analysis are given in Appendix~\ref{sec:coactivation}.

In Section~\ref{sec:application}, we apply the interaction formulation into reverse engineering IOI circuit \citep{ioi} and tracing semantic transformation in the model weights. We validate selected
interactions using attribution patching, behavioral probes, and direct weight interventions.

\section{Experiments}
\label{sec:exp}

We evaluate three claims motivated by Section~\ref{sec:method}: (i) whether the
learned weight components are interpretable and diverse; whether the information written by a component has the same meaning with the component itself, (ii) whether they support localized causal weight editing, and (iii) whether ASPD's internal learning signal improves parameter
decomposition in larger pretrained models.

{\bf Setup.}
We compare ASPD, PD Transcoder, VPD \citep{vpd}, and VPD + internal, which
augments VPD with $\cL_{\mathrm{internal}}$ from
Equation~\ref{eq:internal}. We evaluate one weight matrix in each of three
pretrained models: $MLP_{\mathrm{in}}$ at layer 0 of GPT-2 small
\citep{gpt2}, $MLP_{\mathrm{out}}$ at layer 13 of Gemma-2-2B
\citep{gemma_2}, and the attention $O$ matrix at layer 17 of Qwen-3-8B
\citep{qwen3}. All decompositions are trained on 2B tokens with average
sparsity $L_0=32$, using $24{,}576$, $36{,}864$, and $36{,}864$ weight
components, respectively. For evaluations involving activation features, we
train an independent SAE at the output of the target matrix, so the evaluation
does not use ASPD's own shared representation. Full training details are in
Appendix~\ref{sec:training_details}. We report 95\% confidence intervals.

\subsection{Interpretability, Diversity, and Meaning Localization}
\label{sec:interp}

We evaluate three complementary properties. \textbf{Interpretability:} uses the Intruder
score \citep{intruder}. An LLM judge receives four examples on which a
component fires and one example from another component and must identify the
intruder; a random choice therefore has an accuracy of $0.2$. We evaluate 200 components per method.
\textbf{Diversity:} measures the mean pairwise Jaccard overlap between the tokens on
which two components fire; lower overlap indicates less redundant components.
We evaluate 500 components after filtering extremely sparse or dense
components. Full details are in Appendix~\ref{sec:interp_details}. {\bf Meaning Localization:}
interpretability alone does not establish that the information written by a
weight component has the same meaning as the component itself. To test this, we
use an \textit{independently} trained SAE at the output $y_t$ and estimate which activation feature
each component most strongly affects over $10^6$ held-out tokens via attribution patching \citep{attribution_patching}. We construct 200
component--feature pairs using these estimated effects. We follow the evaluation in
\citet{featflow, semantic_optimal_transport}: an LLM judge compares
activation examples from each component and feature and assigns scores of
3, 2, or 1 for similar, uncertain, or different meanings. We subtract the
score obtained from random component--feature pairings and report the resulting
Matching score. Higher values indicate stronger semantic alignment between a
weight component and the activation feature it affects. This directly
evaluates the coherence between the component activation and what information it writes. Full details are in
Appendix~\ref{sec:localization_details}.

Table~\ref{tab:interp} and \ref{tab:localize} show a clear separation from VPD on the two larger
models. On Gemma-2-2B, ASPD reaches an Intruder score of $0.62$ versus $0.20$
for VPD, while component overlap decreases from $0.44$ to $0.03$. On
Qwen-3-8B, ASPD remains interpretable ($0.57$) and diverse ($0.03$), whereas
VPD is near chance ($0.22$). ASPD also achieves positive Match scores on all
three models, showing that its downstream effects align semantically with
independent activation features. PD Transcoder also performs strongly, while
ASPD is substantially stronger on the Gemma MLP decomposition.

\begin{table}[h]
    \scriptsize
    \centering
    \vspace{-5mm}
    \caption{Interpretability and Diversity experiment results. Interp $=0.2$ means random chance.}
    \begin{tabular}{l|c|c|c|c|c|c}
        \toprule
        & \multicolumn{2}{c|}{GPT2} & \multicolumn{2}{c|}{Gemma-2-2b} & \multicolumn{2}{c}{Qwen-3-8b} \\
        Method & Interp $\uparrow$ & Sim $\downarrow$ & Interp $\uparrow$ & Sim $\downarrow$ & Interp $\uparrow$ & Sim $\downarrow$ \\
        \midrule
        ASPD (Ours) & \textbf{0.68} $\pm$ \textbf{0.04} & \underline{0.01} $\pm$ \underline{0.00} & \textbf{0.62} $\pm$ \textbf{0.03} & \textbf{0.03} $\pm$ \textbf{0.00} & \underline{0.57} $\pm$ \underline{0.04} & \textbf{0.03} $\pm$ \textbf{0.00} \\
        PD Transcoder (Ours) & \underline{0.54} $\pm$ \underline{0.04} & \textbf{0.00} $\pm$ \textbf{0.00} & \underline{0.29} $\pm$ \underline{0.03} & \underline{0.10} $\pm$ \underline{0.01} & \textbf{0.60} $\pm$ \textbf{0.04} & \underline{0.05} $\pm$ \underline{0.00} \\
        VPD & 0.37 $\pm$ 0.03 & \underline{0.01} $\pm$ \underline{0.00} & 0.20 $\pm$ 0.02 & 0.44 $\pm$ 0.01 & 0.22 $\pm$ 0.02 & 0.18 $\pm$ 0.00 \\
        VPD + internal  & 0.43 $\pm$ 0.03 & \textbf{0.00} $\pm$ \textbf{0.00} & 0.25 $\pm$ 0.02 & 0.14 $\pm$ 0.00 & 0.24 $\pm$ 0.02 & 0.24 $\pm$ 0.01 \\
        \bottomrule
    \end{tabular}
    \label{tab:interp}
\end{table}

\begin{table}[h]
    \scriptsize
    \vspace{-5mm}
    \centering
    \caption{Meaning localization experiment results. Matching $=0$ means random chance.}
    \begin{tabular}{l|c|c|c}
        \toprule
        & \multicolumn{1}{c|}{GPT2} & \multicolumn{1}{c|}{Gemma-2-2B} & \multicolumn{1}{c}{Qwen-3-8b} \\
        Method & Matching $\uparrow$ & Matching $\uparrow$ & Matching $\uparrow$ \\
        \midrule
        ASPD (Ours) & \textbf{1.02} $\pm$ \textbf{0.14} & \textbf{0.30} $\pm$ \textbf{0.11} & \underline{0.61} $\pm$ \underline{0.14}  \\
        PD Transcoder (Ours) & \underline{0.94} $\pm$ \underline{0.15} & \underline{0.27} $\pm$ \underline{0.17} & \textbf{0.64} $\pm$ \textbf{0.15} \\
        VPD & 0.31 $\pm$ 0.11 & -0.02 $\pm$ 0.10 & 0.17 $\pm$ 0.11 \\
        VPD + internal  & 0.45 $\pm$ 0.13 & 0.07 $\pm$ 0.12 & 0.10 $\pm$ 0.14 \\
        \bottomrule
    \end{tabular}
    \vspace{-3mm}
    \label{tab:localize}
\end{table}

\subsection{Causal Weight Editing}

\label{sec:weight_edit}

An useful parameter decomposition should allow selected mechanisms to be
modified without broadly perturbing unrelated activation features. We evaluate
this property using the independent output SAE above. 
We evaluate two settings: (1) Single: we sample one target feature and find the top-k components with strongest attribution patching effect \citep{attribution_patching} over a dataset for $k \in \{1, 5, 10, 20, 50\}$. (2) Multiple: we sample a target feature set where the set size is in $\{1, 5, 10, 20, 50\}$; for each target set, we identify the union of the per-feature top-k components for $k \in \{1, 5, 10\}$ via attribution patching.
We then remove all of the selected components directly from the weight matrix and run forward pass on $10^6$ tokens, we repeat this over 50 target sets for both settings. We report the $localization$ which measures the activation change of the target features divided by the change of non-target features, the higher the better. And since absolute $localization$ can depend on the overall strength of the edited
components, we also normalize each score by an edit of the same number of randomly
selected components, resulting in the $ratio$ metric. Having $ratio > 1$ means the edit is better than a random edit. The exact definitions of the scores and full protocol are given in
Appendix~\ref{sec:weight_editing_details}.

Table~\ref{tab:edit} shows that ASPD and PD Transcoder consistently support
localized weight interventions, while VPD is typically at or below random.
The difference is largest on Qwen-3-8B, where ASPD reaches $6.7\times$ the
random baseline for a single target and $11.1\times$ for multi-target edits,
compared with $0.9\times$ and $0.8\times$ for VPD. Thus, the learned weight
components can be manipulated directly with effects that remain concentrated
on the intended activation features.


\begin{table}[h]
    \scriptsize
    \centering
    \vspace{-3mm}
    \caption{Weight Editing Localization experiment results. $ratio < 1$ is poorer than random chance.}
    \begin{tabular}{l|c|c|c|c|c|c}
        \toprule
        & \multicolumn{2}{c|}{GPT2} & \multicolumn{2}{c|}{Gemma-2-2b} & \multicolumn{2}{c}{Qwen-3-8b} \\
        Method & $ratio$ $\uparrow$ & $localization$ $\uparrow$ & $ratio$ $\uparrow$ & $localization$ $\uparrow$ & $ratio$ $\uparrow$ & $localization$ $\uparrow$ \\
        \midrule
        \multicolumn{7}{l}{\textbf{Single}} \\
        \midrule
        ASPD (Ours) & \textbf{3.1} $\pm$ \textbf{0.5} & \underline{0.040} $\pm$ \underline{0.005} & \underline{2.7} $\pm$ \underline{0.8} & \textbf{0.013} $\pm$ \textbf{0.004} & \textbf{6.7} $\pm$ \textbf{0.9} & \textbf{0.053} $\pm$ \textbf{0.006} \\
        PD Transcoder (Ours)  & \underline{3.0} $\pm$ \underline{0.7} & \textbf{0.049} $\pm$ \textbf{0.010} & \textbf{4.3} $\pm$ \textbf{1.1} & \underline{0.012} $\pm$ \underline{0.004} & \underline{2.5} $\pm$ \underline{0.4} & \underline{0.050} $\pm$ \underline{0.005} \\
        VPD & 0.9 $\pm$ 0.2 & 0.017 $\pm$ 0.002 & 0.5 $\pm$ 0.1 & 0.004 $\pm$ 0.001 & 0.9 $\pm$ 0.2 & 0.020 $\pm$ 0.003 \\
        VPD + internal  & 0.3 $\pm$ 0.2 & 0.007 $\pm$ 0.004 & 0.7 $\pm$ 0.2 & 0.006 $\pm$ 0.001 & 1.1 $\pm$ 0.2 & 0.021 $\pm$ 0.003 \\
        \midrule
        \multicolumn{7}{l}{\textbf{Multiple}} \\
        \midrule
        ASPD (Ours) & \textbf{2.8} $\pm$ \textbf{0.2} & \underline{0.022} $\pm$ \underline{0.001} & \underline{1.8} $\pm$ \underline{0.2} & \textbf{0.008} $\pm$ \textbf{0.001} & \textbf{11.1} $\pm$ \textbf{0.6} & \textbf{0.044} $\pm$ \textbf{0.002} \\
        PD Transcoder (Ours) & \underline{2.4} $\pm$ \underline{0.3} & \textbf{0.027} $\pm$ \textbf{0.003} & \textbf{4.0} $\pm$ \textbf{0.4} & \textbf{0.008} $\pm$ \textbf{0.001} & \underline{2.0} $\pm$ \underline{0.1} & \underline{0.039} $\pm$ \underline{0.002} \\
        VPD & 0.9 $\pm$ 0.1 & 0.015 $\pm$ 0.001 & 0.4 $\pm$ 0.0 & \underline{0.004} $\pm$ \underline{0.000} & 0.8 $\pm$ 0.0 & 0.017 $\pm$ 0.001 \\
        VPD + internal  & 0.1 $\pm$ 0.1 & 0.002 $\pm$ 0.001 & 0.5 $\pm$ 0.0 & \underline{0.004} $\pm$ \underline{0.000} & 0.9 $\pm$ 0.1 & 0.015 $\pm$ 0.001 \\
        \bottomrule
    \end{tabular}
    \label{tab:edit}
    \vspace{-3mm}
\end{table}

\subsection{Ablations and Scalability}
\label{sec:exp_ablation}

{\bf Internal learning signal.}
We test our claim of using internal learning signal could improve VPD interpretability in  
Section~\ref{sec:aspd}. As shown in Table~\ref{tab:interp}, adding
$\cL_{\mathrm{internal}}$ improves VPD's interpretability, diversity for all three models. We also report the results of VPD + internal on all metrics in Table \ref{tab:edit} and \ref{tab:localize}. On weight editing, the internal loss yields noticeably improvement on large models but not for GPT2 small, while on meaning localization, it yields improvement on GPT2 small but mixed results on the other two models. These results suggest that the local internal signal can improve VPD's interpretability, but is not sufficient to improve on other metrics.

{\bf Robustness to VPD evaluation choices.}
We test a more favorable setup of VPD on interpretability and diversity. Following \citet{vpd}, we consider VPD components fire iff $g_{t,c} > \tau, \; \tau \in \{0.01, 0.1\}$, which filters spurious activation examples of the components. We found that even in this easier setup, VPD still has Interp score near random chance for large models (Appendix~\ref{sec:act_filter_interp}).

{\bf VPD-specific objectives.}
Finally, we evaluate the $\cL_{\mathrm{param}}, \cL_{\mathrm{ablate}}$ (see Appendix \ref{sec:background}) objectives proposed by previous works \citep{vpd}. Their effects on VPD are
mixed, while adding them to PD Transcoder generally degrades the performance on many metrics. 
We therefore did not include these losses to our solution; the complete ablations are reported in Appendix~\ref{sec:ablate_param}.

\section{Mechanistic Case Studies}
\label{sec:application}

For these analyses, we train ASPD jointly across all 72 weight matrices of
GPT-2 small, with 6,144 components per matrix and 442,368 weight components in
total. This model-wide decomposition allows us to analyze interactions across
matrices, heads, and layers. Training details are provided in
Appendix~\ref{sec:gpt2_decomp}.

\subsection{Recovering Parameter-Level IOI Mechanisms}
\label{sec:ioi_circuit}

Existing circuit analyses \citep{feature_circuit, on_biology, lorsa}
primarily characterize computation in activation space, identifying features
and their interactions but not the underlying weight components that implement
those computations. ASPD provides a complementary parameter-space view by
decomposing the weights themselves into read--write mechanisms. We demonstrate
this capability by recovering parameter-level mechanisms underlying the
classic IOI circuit \citep{ioi}. We highlight the main findings here; the full
analysis is provided in Appendix~\ref{sec:ioi_circuit_full}
(Figures~\ref{fig:qk_h3_0}--\ref{fig:ov_h11_10}).


Figure~\ref{fig:ioi_circuit_mech} summarizes the resulting parameter-level
mechanism circuit. Within individual heads, ASPD recovers components corresponding to known IOI
computations. In induction head H5.5, we found component $164Q$ activates on ``names" and
$5897K$ on ``tokens following names"; removing this pair suppresses the induction
attention pattern. At the output of the same head, ASPD identifies $228O$ as
an \textit{induction-related} component Figure~\ref{fig:ov_h5_5}, we verified this finding independently via testing the component with induction probe (Appendix \ref{sec:probe-induction}). ASPD similarly
identifies $224O$ in duplicate-token head H3.0 as a component associated with
repeated names and tokens.

The read--write interaction analysis further connects these mechanisms across
heads. As shown in Figure~\ref{fig:v_in_h8_6}, the \textit{duplicate-name} component $224O$ of H3.0 and the \textit{induction} component $228O$ of H5.5 are among the strongest contributors to components $2861V$ and $101V$ of the downstream S-inhibition head H8.6. Inspecting deeper, we found that $2861V, 101V$ activate on names and pronouns, \textit{with stronger responses to repeated occurrences} (we verified via a probe Appendix \ref{sec:probe-factorial}), indicating that H8.6 receives both duplicate-token and induction information from upstream heads.

\begin{figure}[h]
    \centering
    \vspace{-3mm}
    \includegraphics[width=0.8\linewidth]{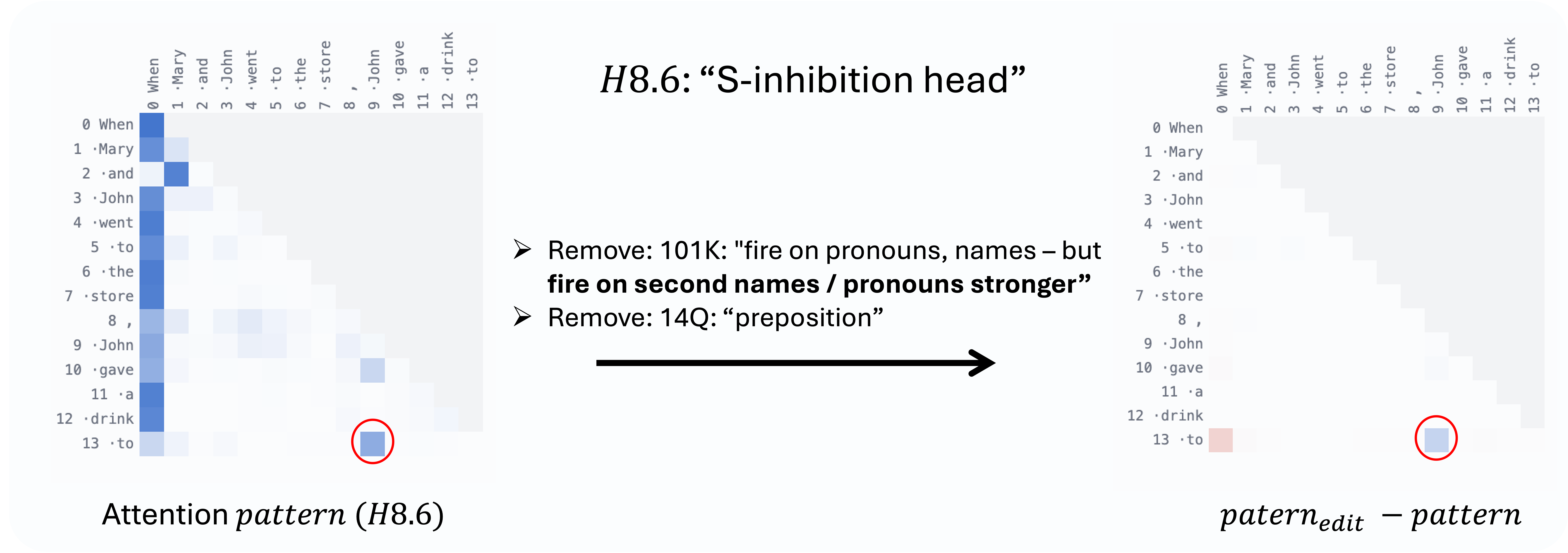}
    \caption{
    Weight editing on the S-inhibition mechanism in H8.6.
    The left panel shows the original attention pattern. The right panel shows
    the change after removing the identified $QK$ component pair from the
    weights. The edit substantially suppresses attention at the
    S-inhibition position.
    }
    \label{fig:qk_h8_6_compact}
\end{figure}

Within H8.6, components $14Q$ and $101K$ (which takes the $224O$ of H3.0 and $228O$ of H5.5 as input) contribute strongly to the
S-inhibition attention pattern. Figure~\ref{fig:qk_h8_6_compact} shows the
effect of directly removing this QK component pair from the frozen weights:
the S-inhibition attention pattern is substantially suppressed. The full
component-level analysis is shown in Figure~\ref{fig:qk_h8_6} in the appendix.
This direct intervention provides causal evidence for the identified
parameter-level mechanism, while attribution patching and behavioral probes
provide supporting evidence.

The recovered mechanism also extends downstream. The overview in
Figure~\ref{fig:ioi_circuit_mech} shows that output components of H8.6 provide
inputs to the query-side components of the name-mover and negative-name-mover
heads, connecting S-inhibition to the later name-selection computation. Thus,
ASPD recovers not only components within individual heads but also
parameter-level connections among the established IOI head classes.

The decomposition does not isolate every mechanism at the finest possible
granularity. For example, H5.5 contains a component representing ``token
following a name" rather than the more specific ``token after John'' signal,
and several similarly behaving components remain only partially understood.
These cases suggest that some mechanisms may require finer decompositions or
additional analysis.

\subsection{Tracing Semantic Transformations Through Model Weights}
\label{sec:knowledge}

Prior work has studied knowledge in model parameters through knowledge editing
\citep{weight_edit, think_weight_edit}, interpretation of MLP weights
\citep{kv_mlp}, and localization of memorized information
\citep{localization_method_localize_memorized_data}. These approaches
typically begin from a known fact, behavior, or parameter representation and
do not jointly characterize how weight-space mechanisms interact with the
activation features they read and write. ASPD provides a complementary,
unsupervised view: its model-wide decomposition allows us to analyze both
component--component interactions, which reveal transformations implemented
by the weights, and component--feature interactions, which reveal how those
transformations read information from and write information back to activation
space. Interaction details are provided in Appendix~\ref{sec:coactivation}. Figure~\ref{fig:knowledge_news} illustrates this analysis for the layer-7 MLP
of GPT-2 small.

\begin{figure}[h]
    \centering
    \includegraphics[width=\linewidth]{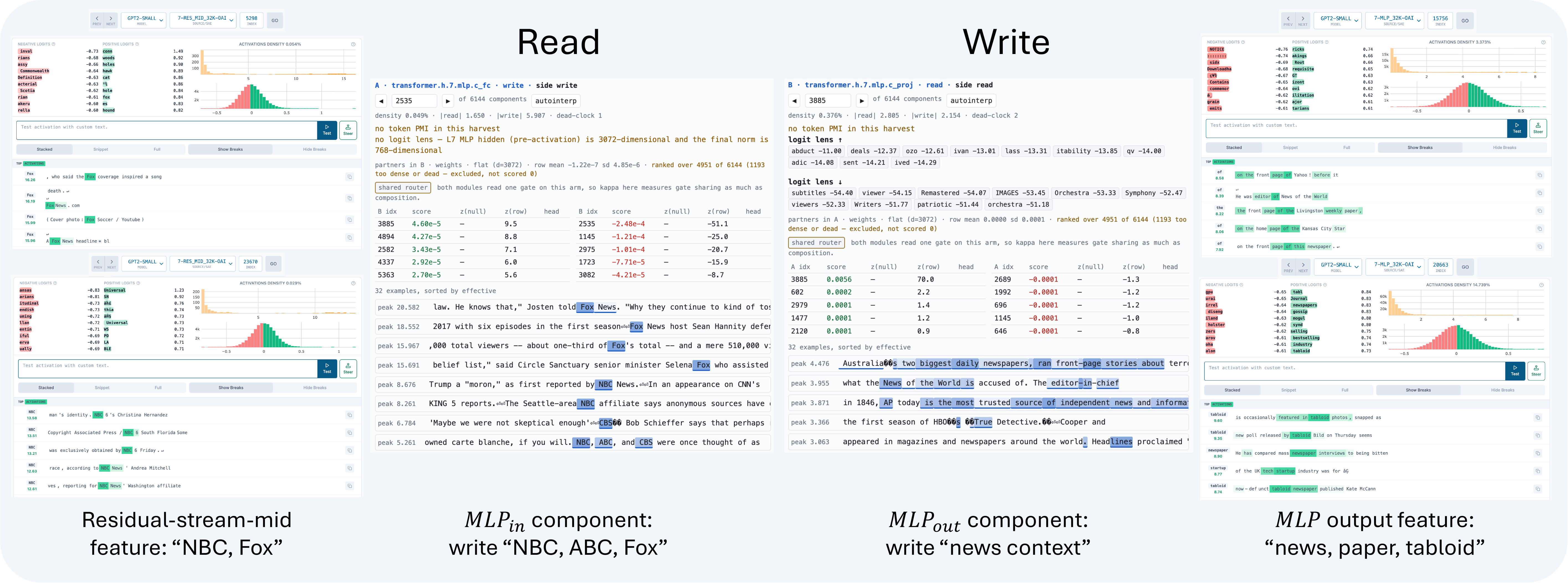}
    \caption{
    Tracing a semantic transformation through the layer-7 MLP.
    $MLP_{\mathrm{in}}$ component 2535, associated with contexts involving
    NBC, ABC, and Fox, reads news-organization features from the residual
    stream and interacts strongly with $MLP_{\mathrm{out}}$ component 3885,
    which writes to downstream features associated with broader news-related
    context.
    }
    \label{fig:knowledge_news}
\end{figure}

As shown in Figure~\ref{fig:knowledge_news}, upstream activation features
associated with news organizations such as NBC and Fox interact strongly with
$MLP_{\mathrm{in}}$ component 2535, which activates on contexts involving
NBC, ABC, Fox, and related entities. Among the $MLP_{\mathrm{out}}$
components, component 3885 has the strongest interaction with component 2535
and activates in broader news-related contexts. Downstream activation features
associated with concepts such as news, papers, and tabloids in turn interact
strongly with component 3885.

This example traces a localized semantic transformation from information about
specific news organizations to a more general news-related representation. It
illustrates how ASPD connects activation features to explicit weight-space
transformations and then back to downstream activation features, characterizing
how semantic information is read, transformed, and written during model
computation. Additional examples are provided in
Figures~\ref{fig:knowledge_tech} and~\ref{fig:knowledge_hack} in the appendix.

\section{Conclusion}
\label{sec:conclusion}
We introduced activation-grounded parameter decomposition, a framework
for interpreting model weights through the activation-space features they
read and write. ASPD provides a scalable realization of this
framework by jointly learning activation features and parameter components,
using an internal reconstruction signal to learn localized mechanisms directly
at the weight matrix being analyzed. Across GPT-2, Gemma-2-2B, and Qwen-3-8B,
ASPD produces interpretable, semantically grounded, and causally editable
parameter components. Moreover, composing their read--write interactions
enables parameter-level mechanism circuits. On GPT-2, ASPD refines the known
IOI circuit to individual weight components and traces how semantic information
flows through model parameters. We hope this perspective provides a foundation
for studying not only what information models represent, but
how their weights transform that information into computation.

\section{AI use statement}
In this work, we used generative AI tools for: 
provide feedback on research methodology, implement methods, support qualitative and thematic data analysis.
We have \textit{not} used generative AI tools for 
assist with translation, interpret results; 
while the following are not applicable to this work: 
help develop theoretical models or conceptual frameworks, formulate mathematical claims, provide critical ingredients for proving mathematical claims, assist in the writing of proofs, propose or refine hypotheses, clean and reformat dataset. 
Additionally, we used AI for polishing the writing and draft the conclusion section of the paper.
We have reviewed all AI-assisted work. We checked the LLM generated code with test runs to ensure we are able to reproduce baseline results and metrics, we checked the writing. We take responsibility for the final content of this work, including text, claims or artifacts produced with the aid of generative AI.






\bibliography{iclr2027_conference}
\bibliographystyle{iclr2027_conference}

\appendix
\section{Background}
\label{sec:background}
The state-of-the-art adVersarial Parameter Decomposition (VPD) \citep{vpd} finds a replacement of $W$ with $C$ rank-1 \emph{components} ($C > \max(\dout,\din)$) $P_c=u_cv_c^\top$, $u_c\in\R^{\dout}$, $v_c\in\R^{\din}$, and a residual $\Delta:=W-\sum_c u_cv_c^\top$, under some constrain $\cL_{VPD}$ loss function.

In mechanistic interpretability, we would like to design the loss $\cL_{VPD}$ so that the solution contains specific properties: interpretable and causally weight-editable. To formalize these properties, existing works \citep{spd, vpd} formulate the decomposition as learning a sparse \emph{causal-importance} function $g(X): \R^{T\times\din} \to [0,1]^{T\times C}$ to control when the components importance. $g_{t,c} > 0$ means that the component $c$ is important in the computation at position $t$, while $g_{t,c} = 0$ means that the component is not important and can be removed from the computation. A component is interpretable if $g_{t,c} > 0$ on a sparse, predictable pattern such as on certain tokens, contexts, or ideas. To formalize causally weight editing, unimportant components are causally editable given \textit{any} combination of ablation masks $m\in\mathcal{M}(X)$ while still preserving the model performance, where $\mathcal M(X)$ is a space of feasible masks on input $X$, $m_{t,c} \in [g_{t,c}(X), 1]$ and $m_{t,\Delta}\in[0,1]$. Concretely, let $\mathrm{D}$ be a divergence metric (such as KL divergence) summed over positions, let $f: \R^{T\times\dout} \to \R^{T\times d_{vocab}}$ map the output activations of component $W$ to the output vocabulary distributions of the model, and let $Y'(m;X)=(y'_1,\dots,y'_T)$ with $y'_t=\sum_c m_{t,c}\,u_c\,(v_c^\top x_t)+m_{t,\Delta}\,\Delta x_t$ be the output activations under the decomposition mask $m$, \citet{vpd} optimize
\begin{align}
\cL_{param}(u, v) &= \Big\|\,W-\textstyle\sum_{c}u_cv_c^\top\Big\|_F^2,
\label{eq:param} \\
\cL_{ablate}(u, v, g) &= \E_{X}\,\E_{m\sim\mu(X)}
\Big[\,\mathrm{D}\big(f(Y)\,\big\|\,f(Y'(m;X))\big)\Big],
\label{eq:ablate}\\
\cL_{sparse}(g) &= \E_{X}\Big[\tfrac{1}{T}\textstyle\sum_t\sum_c \big(1+ \lambda_{freq}\log_2\big(1+\sum_{(X',t')\in\mathcal{B}}|g_{t',c}(X')|^p\big)\big)|g_{t,c}(X)|^p\Big],
\label{eq:imp}\\
\cL_{VPD} &= \lambda_{param} \cL_{param} + \lambda_{ablate} \cL_{ablate} +  \lambda_{sparse} \cL_{\mathrm{sparse}},
\label{eq:all_vpd_loss}
\end{align}
where $\mathcal{B}$ is the set of token positions in a batch, $\mu(X)$ is a distribution over $\mathcal{M}(X)$, and $p$, $\lambda_{param}, \lambda_{ablate}, \lambda_{sparse}, \lambda_{freq}$ are hyperparameters. VPD uses $\mu(X)$ as a mixture of a stochastic mask, $m_{t,c}=g_{t,c}(X)+(1-g_{t,c}(X))\,\xi_{t,c}$ and $m_{t,\Delta}=\xi_{t,\Delta}$ with $\xi_{t,\cdot}\overset{\text{iid}}{\sim}\mathcal{U}[0,1]$, and an adversarial mask, the maximizer of $\mathrm{D}$ over $\mathcal{M}(X)$ approximated by projected gradient ascent. $\cL_{ablate}(u,v,g)$ optimizes the output distribution of the masked computation $f(Y'(m;X))$, under \textit{any} mask combination $m\in\mathcal{M}(X)$, is close to the original model distribution $f(Y)$, complying the editable property. $\cL_{param}(u,v)$ makes sure the decomposition is close to the original weight $W$. $\cL_{sparse}(g)$ forces the importance to be sparse making the decomposition more interpretable.

\paragraph{Limitations.} \citet{vpd} are only able to decompose 4-layers toy model, while in larger models, the decomposition is uninterpretable and has poor editing quality.

\section{Transcoder for Parameter Decomposition}
\label{sec:trans_reform}
In this section, under our formulation (Equation \ref{eq:jointly}), we show that Transcoder \citep{transcoder_circuit} can be recast as a parameter decomposition method, which we denote as \textit{PD Transcoder}. This has the following implications: (1) it unifies both decomposition problems, treating them as similar problems; (2) we can bring many advancements in activation decomposition to parameter decomposition. Given $W^{enc} \in \R^{C \times \din}, W^{dec} \in \R^{C \times \dout}$ be encoder and decoder matrices of the Transcoder. We omit the Transcoder bias for simplicity. We can map our formulation to the Transcoder:

\begin{align}
    \lambda_{act} &= 0\\
    x_t &= r_t \\
    g^s_{t}(R)=g^s(r_t) &= W^{enc} x_t \\
    \phi(g_c^s(r_t)) &= \mathbb{I}[g_c^s(r_t) > 0] \\
    v_c &= W^{enc}_{c} \\
    u_c &= W^{dec}_c.
\end{align}
In this view, the causal importance function takes per-token activation as input. The $W^{enc}$ is shared between importance function $g^s$ and $v_c$, therefore the component activation $v_c^Tx_t$ is also causal important value. The activation loss $\cL_{act} = \|r_t - d(g^s(r_t))\|_2^2$ in Equation \ref{eq:jointly} is not necessary because the Transcoder already reconstruct at output activation $y_t$ via the $\cL_{internal} = \frac{1}{T}\sum_t \big\|y_t - \textstyle\sum_c \phi(g^s_{c}(r_t))\,u_c(v_c^\top x_t)\big\|_2^2$ loss. 

\section{Training Details}
\label{sec:training_details}
We train both activation and parameter decomposition on dataset specified in the Table \ref{tab:dataset}. For both decomposition problems, we train the methods with the number of components/features of $C=24576,36864,36864; F=24576,36864,36864$ for GPT2s, Gemma-2-2b, Qwen-3-8b, respectively.

\begin{table}[h]
    \centering
    \caption{Datasets used for the paper.}
    \begin{tabular}{l|c}
        \toprule
        Model & dataset \\
        \midrule
        GPT2s & apollo-research/Skylion007-openwebtext-tokenizer-gpt2 \citep{skylion}  \\
        Gemma-2-2b & monology/pile-uncopyrighted  \citep{thepile} \\
        Qwen-3-8b & monology/pile-uncopyrighted \citep{thepile}  \\
        \bottomrule
    \end{tabular}
    \label{tab:dataset}
\end{table}

\subsection{Parameter Decomposition}
\label{sec:pd_details}
To make the notation concise, we denote the reconstruction losses from Equation \ref{eq:jointly} as $\cL_{internal} = \frac{1}{T}\sum_t \big\|y_t - \textstyle\sum_c \phi(g^s_{t,c}(R))\,u_c(v_c^\top x_t)\big\|_2^2$ and $\cL_{act} = \|R - d(g^s(R))\|_2^2$. We also denote FVU loss $\text{fvu}(x, x') = \frac{\|x-x'\|_2^2}{\text{var}(x)}$ where $\text{var}(x)$ is the variance of $x$. The full coefficients of the losses are in Table \ref{tab:coef}. We train parameter decomposition with 2B tokens with sparsity on average $L_0=32$. For PD Transcoder and ASPD, the learning rate is $3.10^{-4}$ with Adam optimizer: $\beta_1=0.9$, $\beta_2=0.99$, while for VPD, the learning rate is $5.10^{-5}$ with Adam optimizer: $\beta_1=0.9$, $\beta_2=0.999$ and cosine anneal, faithful to VPD implementation. The full loss coefficients are in Table \ref{tab:coef}, note that the coefficients of the losses are set so that the losses are proportional to each other at any model.

\begin{table}[h]
\centering
\scriptsize
\setlength{\tabcolsep}{4pt}
\renewcommand{\arraystretch}{1.15}
\caption{Training loss coefficients.}
\begin{tabular}{l c c c c c c c c}
\toprule
\textbf{Loss coefficients} & $\lambda_{sparse}$ & $\lambda_{freq}$ & $\lambda_{ablate\_stoch}$ & $\lambda_{ablate\_adv}$ & $\lambda_{param}$ & $\lambda_{internal}$ & $\lambda_{act}$ & $\lambda_{auxiliary}$ \\
\midrule
\multicolumn{9}{l}{\textbf{GPT-2 small}} \\
\midrule
VPD & 1e-5 & 0.5 & 0.5 & 0.5 & 1000 & & & \\
VPD + internal & 1e-5  & 0.5 & 0.5 & 0.5 & 1000 & 0.5 & & \\
VPD + internal + no param & 1e-5 & 0.5 & 0.5 & 0.5 & & 0.5 & & \\
VPD + internal + no ablate & 1e-5 & 0.5 & & & 1000 & 0.5 & & \\
PD Transcoder &  & & & & & 1.0 & & 0.03125 \\
PD Transcoder + param &  & & & & 1000 & 1.0 & & 0.03125 \\
PD Transcoder + ablate &  & & 0.5 & & & 1.0 & & 0.03125 \\
ASPD &  & & & & & 1.0 & 1.0 & 0.03125 \\
\midrule
\multicolumn{9}{l}{\textbf{Gemma-2-2b}} \\
\midrule
VPD & 1e-5 & 0.5 & 0.5 & 0.5 & 1000 & & & \\
VPD + internal & 1e-5 & 0.5 & 0.5 & 0.5 & 1000 & 0.31 & & \\
VPD + internal + no param & 1e-5 & 0.5 & 0.5 & 0.5 & & 0.31 & & \\
VPD + internal + no ablate & 1e-5 & 0.5 & & & 1000 & 0.31 & & \\
PD Transcoder & & & & & & 1.0 & & 0.03125 \\
PD Transcoder + param & & & & & 1000 & 1.0 & & 0.03125 \\
PD Transcoder + ablate &&  & 0.0011 & & & 1.0 & & 0.03125 \\
ASPD & & & & & & 1.0 & 1.0 & 0.03125 \\
\midrule
\multicolumn{9}{l}{\textbf{Qwen-3-8b}} \\
\midrule
VPD & 1e-5 & 0.5 & 0.5 & 0.5 & 1000 & & & \\
VPD + internal & 1e-5 & 0.5 & 0.5 & 0.5 & 1000 & 0.31 & & \\
VPD + internal + no param & 1e-5 & 0.5 & 0.5 & 0.5 & & 0.31 & & \\
VPD + internal + no ablate & 1e-5 & 0.5 & & & 1000 & 0.31 & & \\
PD Transcoder & & & & & & 1.0 & & 0.03125 \\
PD Transcoder + param & & & & & 1000 & 1.0 & & 0.03125 \\
PD Transcoder + ablate & & & 1.4 & & & 1.0 & & 0.03125 \\
ASPD & & & & & & 1.0 & 1.0 & 0.03125 \\
\bottomrule
\end{tabular}
\label{tab:coef}
\end{table}

\textbf{ASPD:} We use BatchTopK function \citep{batchtopk} for the sparsity function. We use a per-token causal importance function for simplicity: $g^s_{t}(R)=g^s(r_t)$ with $g^s:\R^{d_{act}}\to\R^C$ and we choose $\phi$ as the indicator function $\mathbb{I}[g^s_c(r_t) > 0]$. We use the FVU variant of $\cL_{internal}$ and $\cL_{act}$. 
For the $\cL_{act}$, we also implement Matryoshka loss \citep{matryoshka} with prefix ratio of $[0.0625, 0.0625, 0.125, 0.25, 0.5]$. 
We also train an auxiliary loss as in \citet{topk} to revive dead components with $top\_k\_auxk=512,4608,2048$ for GPT2s, Gemma-2-2b, and Qwen-3-8b, respectively. The input activation $r$ of the shared gate $g^s_c(r_t)$ is the residual stream activation of the model, either before or after the component. For the matrices $Q, K, V$ of Attention and all matrices in the MLP module, we use the $r$ as the residual stream before the weight matrices (i.e. residual-stream-pre for Attention weight matrices, residual stream mid for MLP weights). On the other hand, for matrix $O$ of Attention, $r$ is the residual stream after the Attention (residual-stream-mid). This is because the input residual stream is aggregated via attention patterns before applying matrix $O$; therefore, aligning with the input residual stream would not capture the attention-pattern relationship. Note that the MLP importance function can be trained on either the output or input residual stream, each with its own interpretation: how the component acts given an input feature, or how the components write each feature to the output stream; however, we did not systematically investigate which option is better.

\textbf{VPD:} We follow the training of \citet{vpd}, all notations of VPD are in Appendix \ref{sec:background}. \textbf{Causal importance function:} For experiments in Section \ref{sec:interp}, \ref{sec:weight_edit}, we train the Transformer causal importance function with (1) for GPT2s: hidden state $d_{model}=512$, MLP hidden state $d_{mlp}=2048$, number of attention heads $n_{head}=8$, and number of transformer blocks $n_{block}=4$; for Gemma-2-2b and Qwen-3-8b: hidden state $d_{model}=1024$, MLP hidden state $d_{mlp}=4096$, number of attention heads $n_{head}=8$, and number of transformer blocks $n_{block}=5$. For the experiment in Section \ref{sec:application}, we train one transformer as a causal importance function for the whole GPT2s model, faithful to VPD: hidden state $d_{model}=2048$, MLP hidden state $d_{mlp}=8192$, number of attention heads $n_{head}=16$, and number of transformer blocks $n_{block}=8$. The transformer can see the full context (not autoregressively masked) of activations of all layers; more details are in the VPD code \citep{vpd}. \textbf{Losses:} we train VPD with stochastic loss $\cL_{ablate\_stoch}$ which is $\cL_{ablate}$ with a random mask $m_{t,c}=g_{t,c}(X)+(1-g_{t,c}(X))\,\xi_{t,c}$ and $m_{t,\Delta}=\xi_{t,\Delta}$ where $\xi_{t,\cdot}\overset{\text{iid}}{\sim}\mathcal{U}[0,1]$, and adversarial loss $\cL_{ablate\_adv}$ where the mask is $m = \arg\text{max}_{m \in \mathcal M(X)}\big[\text{D}(f(Y)||f(Y'(m;X))\big]$ for a given $X$. For adversarial loss, VPD finds $m$ by running gradient ascent on the divergence loss $\text{D}(f(Y)||f(Y'(m;X))$; we use the Adam optimizer with $lr=10^{-2}$, $\beta_1=0.5, \beta_2=0.99, eps=10^{-8}$ and 3 optimization steps for each training step to find $m$. For $\cL_{sparse}$ loss, we use $p=2$ and anneal to $p=0.4$ over the training. 

\textbf{Sparsity Adaptive Loss:} Note that we only apply sparsity adaptive loss for all VPD with internal loss variants but \textit{not} for the original VPD to keep the implementation faithful; however, we do compare VPD with and without this adaptive loss in Appendix \ref{sec:compare_sparsity_adaptive} and the results do not change our claims. The reason for this loss is that we found that VPD sparsity loss is often either too weak or too strong in enforcing sparsity, leading to either overly densely activated components or overly sparse components. We therefore monitor the sparsity loss (described in \citet{matryoshka}) via an adaptive rescaling to make sure the sparsity is comparable with ASPD and PD Transcoder. Concretely, let $L_0 = \frac{1}{T}\sum_c\sum_t\mathbb{I}[g_{t,c}(X) > \tau]$ be the effective number of activation per token and $L_0^*$ be the target sparsity level we want to assert, we multiply the coefficient of the sparsity loss with an adaptive at training step $i$: $\log \mu_{i+1} \leftarrow clip(\log \mu_{i-1} + 3.10^{-4}\kappa(e)\tanh(10 e), 0.001, 1000)$ where $e = log\frac{max(\hat L_0,10^{-6})}{L_0^*}$, $\hat L_0$ is smoothing $L_0$ over training ($\hat L_0 \leftarrow 0.99 * \hat L_0 + L_0$, and $\kappa(e) = \begin{cases} 1 & e > 0 \quad \text{(tighten: } L_0 \text{ above target)}\\[2pt]
3 & e \le 0 \quad \text{(loosen)}\end{cases}$, the update is skipped when $|e| < \log 1.15$. This formulation will increase the sparsity loss if the $L_0 > L_0^*$ and vice versa. We apply this adaptive coefficient $\mu_i$ only after 10\% of the training. Please refer to \citet{matryoshka} for implementation.

\textbf{PD Transcoder:} Similar to ASPD, we also use the BatchTopK sparsity function. The decision of $\phi, r_t$ and other formulation details are in Appendix \ref{sec:trans_reform}. We train PD Transcoder with Matryoshka reconstruction loss \citep{matryoshka} for $\cL_{internal}$ with prefix ratio of $[0.0625, 0.0625, 0.125, 0.25, 0.5]$ with auxiliary loss \citep{topk} to revive dead components. Note that, for PD Transcoder with $\cL_{ablate}$ (see VPD implementation above), we adapt the stochastic loss $\cL_{ablate\_stoch}$ by randomly sampling $m_c \in \{\mathbb{I}[g^s_c(x_t) > 0], 1\}$ (we randomly sample $m_c$ from \{0,1\} if $g^s_c(x_t) \leq 0$, else $m_c=1$). We cannot adapt $\cL_{ablate\_adv}$ for PD Transcoder without meaningfully changing the architecture. 

\subsection{Sparse Autoencoder Training}
\label{sec:sae_details}
For all three SAEs used in the three models, we train on the output activation $y_t$ of the components; the SAEs are used in the Weight Editing (Section \ref{sec:weight_edit}) and Meaning Localization (Section \ref{sec:interp}). We train SAEs with 500M tokens with the BatchTopK activation function \citep{batchtopk}, $L_0=32$, and a learning rate of $3.10^{-4}$. We implement Matryoshka reconstruction loss with prefixes ratio of [0.0625, 0.0625, 0.125, 0.25, 0.5], auxiliary loss \citep{topk} with coefficient $0.03125$. We use the Adam optimizer with $\beta_1=0.9, \beta_2=0.99$.

\subsection{GPT2s Decomposition Training}
\label{sec:gpt2_decomp}

In the Application section \ref{sec:application}, we train ASPD for every weight matrix of GPT2s \citep{gpt2} jointly. We decompose all $72$ matrices including $Q$, $K$, $V$, $O$, and both MLP matrices with $C = 8\cdot\min(d_{in},d_{out}) = 6144$ components per matrix, i.e. $72\times 6144 = 442{,}368$
components in total. Everything not restated below is as in Section~\ref{sec:pd_details}: we use the same
$2$B training tokens (OpenWebText at sequence length $512$, batch size $16$, $250$k steps, bf16,
data-parallel over 4 GPUs), the same optimizers and learning rates per method, and the loss
coefficients of Table~\ref{tab:coef}.
All numbers we report come from the final checkpoint. We enforce the sparsity using BatchTopK with $k=32$ at every matrix. Since the shared gate reads the residual stream at the site prescribed in Section~\ref{sec:pd_details}, we train the decomposition of the matrices $Q$, $K$, $V$ using a shared causal importance function that takes input from residual-stream-pre (the residual stream at the input of the attention head), while we train the decomposition of $O$, $MLP_{in}$, $MLP_{out}$ using the residual-stream-mid (the residual stream at the input of the MLP head). This way, we make our implementation lighter without sacrificing performance.

\subsection{Interpretability and Diversity Details}
\label{sec:interp_details}

We say a component $c$ fires at position $t$ \emph{fires} iff $g_{t, c}(X) > 0$. Both metrics use $400$ fired examples collected during the forward pass over batches of $16$ sequences of $512$ tokens ($3.28$M tokens). Each example is $41$ tokens, centered on a firing. We write $\rho_c$ as the fire density of a component. A component is eligible if it has at least $5$ examples.

\textbf{Interpretability (intruder detection).} We follow the intruder protocol \citep{intruder}. For component $c$ we run $\mathcal I=10$ trials; each trial shows the judge $n_{real}=4$ examples of $c$, plus one example of a different component (intruder) $d \neq c$, which is sampled randomly with $|\rho_d - \rho_c| \le 0.05$. The intruder example is inserted at a uniformly random position
$p_i \in \{1,\dots,n_{real}+1\}$.
The judge gives the prediction $\hat\jmath$ of the index of intruder example, we calculate the interpretability score as
\begin{align}
    interp_c = \frac{1}{\mathcal I}\sum_{i=1}^{\mathcal I} \mathbb{I}\big[\hat\jmath_i = p_i\big],
    \qquad
    \overline{interp} = \frac{1}{|K|}\sum_{c \in K} interp_c ,
\end{align}
so chance is $1/(n_{real}+1) = 0.2$. We score a set $K$ of $200$ components for each method. We use Llama-3.3-70B-Instruct \citep{llama3_3} as the judge. 

\textbf{Diversity.} The intruder score cannot measure the redundancy (diversity) of the component set. This is important because a decomposition can have high interpretability score while every component means the same thing. Let $\mathcal T_{c_1}$ be the set of
token types component $c_1$ fires on at least twice across its examples. We measure the mean pairwise overlap
\begin{align}
    sim_{c_1, c_2} = \frac{|\mathcal T_{c_1} \cap \mathcal T_{c_2}|}{|\mathcal T_{c_1} \cup \mathcal T_{c_2}|},
    \qquad
    \overline{sim} = \frac{1}{n(n-1)}\sum_{c_1 \neq c_2} sim_{c_1, c_2} ,
\end{align}
Having score $\overline{sim} = 0$ means the components fire on disjoint token sets and $\overline{sim} = 1$ means they are indistinguishable, the lower the score the better. We restrict to components in the density to $\rho_c \in [5\cdot10^{-5}, 10^{-3}]$ to avoid overly sparsely or densely fired components and estimate $\overline{sim}$ on a uniform sample of $500$ components.

\subsection{Weight Editing Localization Details}
\label{sec:weight_editing_details}

We ask whether we can use parameter decomposition to causally edit the weight of language model. Specifically, we would want the edit to change a picked set of mechanisms while leaving other mechanisms unchanged. We first train a SAE (Appendix \ref{sec:sae_details}) at output activation of a weight matrix $y_t$, then select a set of output features from the SAE (step \textit{Targets} below), select a set of components for each feature and edit all components at once (step \textit{Ranking and edit}), and lastly, measure the effects of the edit (step \textit{Localization} and \textit{Ratio over random}). For all steps, we conduct on a dataset of $10^{6}$ tokens draw from a general corpus.

\textbf{Targets.} We want to perform editing on a combination of components across many target features. Eligible features are $E = \{j : 0 < \rho_j < 0.2 \text{ and } |A_j| \ge 100\}$,
where $A_j$ is the set of tokens at which feature $j$ is active. We draw a pool of $600$ features
from $E$ that is reused by all parameter decomposition methods for evaluation to ensure comparable results. We evaluate 2 settings. (1) Single: we sample 50 features from $E$ and edit each on its own, (2) Multiple: we draw target feature sets J with $|J| \in \{1, 5, 10, 20, 50\}$ from the 600-feature pool, 50 sets per cardinality.

\textbf{Ranking and edit.} Denote $M_{j,c} = \langle W^{enc}_{:,j}, u_c \rangle$ the attribution of component $c$ to feature $j$ where $W^{enc}$ is the encoder of the output SAE. We average the attribution over the tokens where $j$ is active:
\begin{align}
    \text{effect}_{j,c} = M_{j,c}\; \mathbb{E}_{t \in A_j}\big[\zeta_c(t)\big],
    \qquad \zeta_c(t) = g_{t,c}\, v_c^\top x_t ,
\end{align}
The causal $\text{effect}_{j, c}$ computes how ablating component $c$ will affect the activation of feature $j$. We select the component set by taking the union of the per-feature top-$k$ components:
$\text{Comp}(J,k) = \bigcup_{j \in J} \text{top-}k\,(|\text{effect}_{j,\cdot}|)$  with $k \in \{1, 5, 10, 20, 50\}$ for Single and $k \in \{1, 5, 10\}$ for Multiple (up to 50 when $|J|=1$), and delete them from the frozen weight:
\begin{align}
    W'(J, k) = W - \sum_{c \in \text{Comp}(J, k)} u_c v_c^T .
\end{align}
Each score is averaged over $k$ (Single), and over $k$ and then over $|J|$ (Multiple).

\textbf{Localization.} Given the edit, we rerun the model on a dataset of tokens $10^6$ tokens and measure the change of activation of the target feature set $J$ and features not in set the set $J$. We write $\Delta f_i = f_i(y'_t) - f_i(y_t)$ for the change in feature $i$ over
the tokens $A_J = \bigcup_{j \in J} A_j$,
\begin{align}
    localization
    = \frac{\sum_{j \in J} \mathbb{E}_{t\in A_J}\big[|\Delta f_j|\big]}
           {\sum_{j \in J} \mathbb{E}_{t\in A_J}\big[|\Delta f_j|\big]
            + \sum_{i \notin J} \mathbb{E}_{t\in A_J}\big[|\Delta f_i|\big]} \in [0,1].
\end{align}
This value measure how the editing causal effect are localized onto target feature set $J$: $1$ means the edit moved only features in $J$, and a value near $0$ means the same deletion disturbed the rest of the dictionary features. We want localization to be high.

\textbf{Ratio over random.} Localization alone is not comparable across methods, because higher component norm means the edits are stronger. We therefore also report localization
in divide over the localization of a \emph{random} edits (edit random components) given the same setup $ratio = \frac{localization}{localization\_random}$. A ratio of $1$ means the editing effect equals to random edit, the ratio greater than $1$ means the effects are better than random and vice versa.

\subsection{Meaning Localization Details}
\label{sec:localization_details}

While weight editing can be localized, the causal effect of the components can target unrelated features. We therefore evaluate whether the meaning of the components are localized. Specifically, we pair (\textit{Pairing} step) each component with one output SAE feature by finding the feature affected the most by the component. We then have a judge compare their activating examples (\textit{Judging} step) to see how coherence the activating examples of the pair. The more coherence, the more interpretable and localized the causal effect. We follow the judge scheme from \citet{featflow, semantic_optimal_transport}.

\textbf{Pairing.} Let $A_j$ be the set of tokens at which feature $j$ activates and $M_{j,c} = \langle W^{enc}_{:,j}, u_c \rangle$ the attribution of component $c$ to feature $j$ where $W^{enc}$ is the encoder of the output SAE. Let $\zeta_c(t) = g_{t,c}\, v_c^\top x_t$, we compute the causal effect between component $c$ and feature $j$, accumulated over $10^6$ held-out tokens:
\begin{align}
    \text{effect}_{j, c} = \mathbb{E}_{t \in A_j}\big[|\zeta_c(t)|\big]\; M_{j,c},
    \qquad \pi(c) = \arg\max_j (\text{effect}_{j, c}) ,
\end{align}
The causal $\text{effect}_{j, c}$ approximates how ablating component $c$ will affect the activation of feature $j$. We choose the pair $c, j$ by select the features that have the highest causal effect. We reconstruct the pairs to components and features with at least $10$ examples in the harvest, and $200$ pairs are sampled for judging.

\textbf{Judging.} For each pair, the judge is shown activating examples of both components and features, we select random 9 activation examples each. The judge is then look at the examples and output one of three judgement:
\textsc{similar} $\to 3 \; \text{points}$, \textsc{maybe} $\to 2 \; \text{points}$, \textsc{different} $\to 1 \; \text{points}$. The score is the mean over pairs, $\bar S \in [1,3]$. We use the prompt provided in \citet{semantic_optimal_transport} and use Llama-3.3-70B-Instruct \citep{llama3_3} as the judge. 
We report $\bar S - \bar S_{random}$ where $\bar S_{random}$ is the score when measure on random component-feature pairs. This is because some methods produce ambiguous / less interpretable components by default, making the judge not sure if the meanings of the feature and component match or not; in those cases, even the random pairing reach the same score as pairing via causal effect. We therefore report the score margin over random pairings.

\section{Are $\cL_{ablate}, \cL_{param}$ necessary?}
\label{sec:ablate_param}
In this section, we analyze whether adding the VPD proposed losses $\cL_{ablate}, \cL_{param}$ (see Appendix \ref{sec:background}) yields stronger results on the metrics. We use VPD with internal reconstruction (because adding the internal loss improves the VPD performance noticeably, Section \ref{sec:exp}) and PD Transcoder as the baseline; we then either add or remove one of $\cL_{ablate}, \cL_{param}$ and compare with the baseline. The adaptation of the losses on PD Transcoder is given in Appendix \ref{sec:training_details}.

All the results are in Tables \ref{tab:interp_loss}, \ref{tab:edit_loss}, \ref{tab:localize_loss} for each of the metrics, respectively. Adding the two losses noticeably degrades the performance of PD Transcoder on all three metrics. On the other hand, the effect is more complex for VPD runs. We observed that removing the losses makes VPD more interpretable on Gemma-2-2b and Qwen-3-8b but slightly less diverse. Furthermore, the VPD runs without $\cL_{ablate}, \cL_{param}$ losses are less weight-editable compared to the baseline, and the effect is not clear in the Meaning Localization metric. Given the results, we did not add the losses into ASPD or PD Transcoder due to the damage to PD Transcoder performance and the noisy results from VPD runs. 

\textbf{Discussion:} Although the poor result of the $\cL_{ablate}, \cL_{param}$ loss on PD Transcoder, we believe that $\cL_{ablate}$ still provides important properties. The loss forces the decomposition to learn ``modular'' components in the sense that we can edit the components independently in any combination, which could improve the weight editability of the mechanisms. Furthermore, as discussed in \citet{vpd}, it prevents feature splitting \citep{absorption} from occurring by design. However, whether the ``modular components '' are obtainable or whether it meaningfully improves the parameter decomposition in any other way (beyond being more weight-editable \textit{in the ideal case}) is not clear to us; and adapting this loss into PD Transcoder or ASPD requires non-trivial effort; we leave this for future work. 

\begin{table}[h]
    \scriptsize
    \centering
    \caption{Interpretability and Diversity experiment results of adding or removing $\cL_{param}, \cL_{ablate}$.}
    \begin{tabular}{l|c|c|c|c|c|c}
        \toprule
        & \multicolumn{2}{c|}{GPT2} & \multicolumn{2}{c|}{Gemma-2-2b} & \multicolumn{2}{c}{Qwen-3-8b} \\
        Method & Interp $\uparrow$ & Sim $\downarrow$ & Interp $\uparrow$ & Sim $\downarrow$ & Interp $\uparrow$ & Sim $\downarrow$ \\
        \midrule
        PD Transcoder (Ours) & \textbf{0.54} $\pm$ \textbf{0.04} & 0.00 $\pm$ 0.00 & \underline{0.29} $\pm$ \underline{0.03} & \underline{0.10} $\pm$ \underline{0.01} & \textbf{0.60} $\pm$ \textbf{0.04} & \textbf{0.05} $\pm$ \textbf{0.00} \\
        PD Transcoder + param & \underline{0.45} $\pm$ \underline{0.04} & 0.00 $\pm$ 0.00 & 0.24 $\pm$ 0.02 & \textbf{0.07} $\pm$ \underline{0.01} & \underline{0.36} $\pm$ \underline{0.03} & \underline{0.07} $\pm$ \underline{0.00} \\
        PD Transcoder + ablate & 0.18 $\pm$ 0.03 & 0.00 $\pm$ 0.00 & 0.24 $\pm$ 0.02 & 0.13 $\pm$ 0.00 & 0.22 $\pm$ 0.02 & 0.12 $\pm$ 0.00 \\
        VPD + internal  & 0.43 $\pm$ 0.03 & 0.00 $\pm$ 0.00 & 0.25 $\pm$ 0.02 & 0.14 $\pm$ 0.00 & 0.24 $\pm$ 0.02 & 0.24 $\pm$ 0.01 \\
        VPD + internal + no param  & 0.42 $\pm$ 0.03 & 0.00 $\pm$ 0.00 & 0.24 $\pm$ 0.02 & 0.16 $\pm$ 0.00 & 0.28 $\pm$ 0.02 & 0.21 $\pm$ 0.00 \\
        VPD + internal + no ablate & 0.28 $\pm$ 0.03 & 0.00 $\pm$ 0.00 & \textbf{0.30} $\pm$ \textbf{0.03} & 0.16 $\pm$ 0.00 & 0.27 $\pm$ 0.02 & 0.16 $\pm$ 0.00 \\
        \bottomrule
    \end{tabular}
    \label{tab:interp_loss}
\end{table}

\begin{table}[h]
    \scriptsize
    \centering
    \caption{Weight Editing Localization experiment results of adding or removing $\cL_{param}, \cL_{ablate}$.}
    \begin{tabular}{l|c|c|c|c|c|c}
        \toprule
        & \multicolumn{2}{c|}{GPT2} & \multicolumn{2}{c|}{Gemma-2-2b} & \multicolumn{2}{c}{Qwen-3-8b} \\
        Method & $ratio$ $\uparrow$ & $localization$ $\uparrow$ & $ratio$ $\uparrow$ & $localization$ $\uparrow$ & $ratio$ $\uparrow$ & $localization$ $\uparrow$ \\
        \midrule
        \multicolumn{7}{l}{\textbf{Single}} \\
        \midrule
        PD Transcoder (Ours)  & \textbf{3.0} $\pm$ \textbf{0.7} & \textbf{0.049} $\pm$ \textbf{0.010} & \textbf{4.3} $\pm$ \textbf{1.1} & \underline{0.012} $\pm$ \underline{0.004} & \underline{2.5} $\pm$ \underline{0.4} & \underline{0.050} $\pm$ \underline{0.005} \\
        PD Transcoder + param & \underline{1.6} $\pm$ \underline{0.034} & \underline{0.046} $\pm$ \underline{0.008} & 2.2 $\pm$ 0.7 & \textbf{0.013} $\pm$ \textbf{0.004} & \textbf{2.6} $\pm$ \textbf{0.4} & \underline{0.047} $\pm$ \underline{0.006} \\
        PD Transcoder + ablate & 0.6 $\pm$ 0.2 & 0.016 $\pm$ 0.004 & \underline{2.6} $\pm$ \underline{0.6} & \underline{0.012} $\pm$ \underline{0.003} & 1.0 $\pm$ 0.2 & 0.018 $\pm$ 0.003 \\
        VPD + internal  & 0.3 $\pm$ 0.2 & 0.007 $\pm$ 0.004 & 0.7 $\pm$ 0.2 & 0.006 $\pm$ 0.001 & 1.1 $\pm$ 0.2 & 0.021 $\pm$ 0.003 \\
        VPD + internal + no param  & 0.3 $\pm$ 0.2 & 0.006 $\pm$ 0.004 & 0.0 $\pm$ 0.0 & 0.000 $\pm$ 0.000 & 1.1 $\pm$ 0.3 & 0.020 $\pm$ 0.003 \\
        VPD + internal + no ablate & 0.1 $\pm$ 0.1 & 0.002 $\pm$ 0.002 & 0.4 $\pm$ 0.1 & 0.001 $\pm$ 0.000 & 0.4 $\pm$ 0.2 & 0.001 $\pm$ 0.001 \\
        \midrule
        \multicolumn{7}{l}{\textbf{Multiple}} \\
        \midrule
        PD Transcoder (Ours) & \textbf{2.4} $\pm$ \textbf{0.3} & \textbf{0.027} $\pm$ \textbf{0.003} & \textbf{4.0} $\pm$ \textbf{0.4} & \textbf{0.008} $\pm$\textbf{ 0.001} & \textbf{2.0} $\pm$ \textbf{0.1} & \textbf{0.039} $\pm$ \textbf{0.002} \\
        PD Transcoder + param & \underline{1.2} $\pm$ \underline{0.2} & \underline{0.021} $\pm$ \underline{0.003} & 1.5 $\pm$ 0.2 & \textbf{0.008} $\pm$ \textbf{0.001} & \textbf{2.0} $\pm$ \textbf{0.1} & \underline{0.035} $\pm$ \underline{0.002} \\
        PD Transcoder + ablate & 0.7 $\pm$ 0.1 & 0.014 $\pm$ 0.001 & \underline{2.2} $\pm$ \underline{0.2} & \textbf{0.008} $\pm$ \textbf{0.001} & 0.7 $\pm$ 0.0 & 0.012 $\pm$ 0.001 \\
        VPD + internal  & 0.1 $\pm$ 0.1 & 0.002 $\pm$ 0.001 & 0.5 $\pm$ 0.0 & \underline{0.004} $\pm$ \underline{0.000} & 0.9 $\pm$ 0.1 & 0.015 $\pm$ 0.001 \\
        VPD + internal + no param  & 0.1 $\pm$ 0.0 & 0.002 $\pm$ 0.001 & 0.2 $\pm$ 0.0 & 0.000 $\pm$ 0.000 & 0.9 $\pm$ 0.1 & 0.015 $\pm$ 0.001 \\
        VPD + internal + no ablate & 0.1 $\pm$ 0.0 & 0.001 $\pm$ 0.000 & 0.5 $\pm$ 0.0 & 0.001 $\pm$ 0.000 & 0.7 $\pm$ 0.1 & 0.001 $\pm$ 0.000 \\
        \bottomrule
    \end{tabular}
    \label{tab:edit_loss}
\end{table}

\begin{table}[h]
    \footnotesize
    \centering
    \caption{Meaning Localization results of adding or removing $\cL_{param}, \cL_{ablate}$.}
    \begin{tabular}{l|c|c|c}
        \toprule
        & \multicolumn{1}{c|}{GPT2} & \multicolumn{1}{c|}{Gemma-2-2B} & \multicolumn{1}{c}{Qwen-3-8b} \\
        Method & Matching $\uparrow$ & Matching $\uparrow$ & Matching $\uparrow$ \\
        \midrule
        PD Transcoder (Ours) & \underline{0.94} $\pm$ \underline{0.15} & \underline{0.27} $\pm$ \underline{0.17} & \textbf{0.64} $\pm$ \textbf{0.15} \\
        PD Transcoder + param & \textbf{1.14} $\pm$ \textbf{0.14} & 0.17 $\pm$ 0.15 & \underline{0.22} $\pm$ \underline{0.15}  \\
        PD Transcoder + ablate & 0.33 $\pm$ 0.16 & -0.01 $\pm$ 0.14 & 0.12 $\pm$ 0.14 \\
        VPD + internal  & 0.45 $\pm$ 0.13 & 0.07 $\pm$ 0.12 & 0.10 $\pm$ 0.14 \\
        VPD + internal + no param  & 0.72 $\pm$ 0.14 & -0.06 $\pm$ 0.14 & 0.20 $\pm$ 0.16 \\
        VPD + internal + no ablate & 0.47 $\pm$ 0.15 & \textbf{0.33 }$\pm$ \textbf{0.15} & 0.16 $\pm$ 0.17 \\
        \bottomrule
    \end{tabular}
    \label{tab:localize_loss}
\end{table}

\section{Additional Interpretability and Diversity Results with Activation Filtering}
\label{sec:act_filter_interp}

In this section, we evaluate Interpretability and Diversity as in Section \ref{sec:interp} but with an easier setup that favors the VPD method more: we follow \citet{vpd} to filter the tokens with low causal importance value $g_{t,c} > \tau$ where $\tau \in \{0.01, 0.1\}$. This procedure was measured in \citet{vpd} and was observed to improve the interpretability score of VPD. The remaining setups are the same as described in Appendix \ref{sec:interp_details}. The results are in Table \ref{tab:interp_filter}. We found that even with this filtering, VPD and VPD with internal reconstruction loss still could not produce interpretable components on larger model as in Gemma-2-2b or Qwen-3-8b. 

\begin{table}[h]
    \scriptsize
    \centering
    \caption{Interpretability and Diversity experiment results of filtering $g_{t,c} > \tau$.}
    \begin{tabular}{l|c|c|c|c|c|c}
        \toprule
        & \multicolumn{2}{c|}{GPT2} & \multicolumn{2}{c|}{Gemma-2-2b} & \multicolumn{2}{c}{Qwen-3-8b} \\
        Method & Interp $\uparrow$ & Sim $\downarrow$ & Interp $\uparrow$ & Sim $\downarrow$ & Interp $\uparrow$ & Sim $\downarrow$ \\
        \midrule
        ASPD (Ours) & \textbf{0.68} $\pm$ \textbf{0.04} & \underline{0.01} $\pm$ \underline{0.00} & \textbf{0.62} $\pm$ \textbf{0.03} & \textbf{0.03} $\pm$ \textbf{0.00} & \underline{0.57} $\pm$ \underline{0.04} & \textbf{0.03} $\pm$ \textbf{0.00} \\
        PD Transcoder (Ours) & \underline{0.54} $\pm$ \underline{0.04} & \textbf{0.00} $\pm$ \textbf{0.00} & 0.29 $\pm$ 0.03 & 0.07 $\pm$ 0.01 & \textbf{0.60} $\pm$ \textbf{0.04} & \underline{0.05} $\pm$ \underline{0.00} \\
        VPD + ($g_{t,c}>0.01$)  & 0.41 $\pm$ 0.03 & \textbf{0.00} $\pm$ \textbf{0.00} & 0.22 $\pm$ 0.02 & 0.46 $\pm$ 0.01 & 0.24 $\pm$ 0.02 & 0.19 $\pm$ 0.02 \\
        VPD + internal + ($g_{t,c}>0.01$)  & 0.36 $\pm$ 0.03 & \textbf{0.00} $\pm$ \textbf{0.00} & 0.32 $\pm$ 0.03 & 0.08 $\pm$ 0.01 & 0.26 $\pm$ 0.02 & 0.13 $\pm$ 0.02 \\
        VPD + ($g_{t,c}>0.1$)  & 0.53 $\pm$ 0.03 & \textbf{0.00} $\pm$ \textbf{0.00} & 0.22 $\pm$ 0.02 & 0.46 $\pm$ 0.01 & 0.26 $\pm$ 0.02 & 0.22 $\pm$ 0.01 \\
        VPD + internal + ($g_{t,c}>0.1$) & 0.41 $\pm$ 0.03 & \textbf{0.00} $\pm$ \textbf{0.00} & \underline{0.38} $\pm$ \underline{0.03} & \underline{0.04} $\pm$ \underline{0.01} & 0.34 $\pm$ 0.03 & 0.15 $\pm$ 0.02 \\
        \bottomrule
    \end{tabular}
    \label{tab:interp_filter}
\end{table}

\section{Compare VPD with and without Sparsity Adaptive Loss}
\label{sec:compare_sparsity_adaptive}\textbf{}

In the experiment in Section \ref{sec:exp} and Appendix \ref{sec:ablate_param}, we add the sparsity-adaptive loss described in Appendix \ref{sec:training_details} to all of the VPD internal loss variants because we observed that VPD runs often learn highly dense components that would not be interpretable; however, we keep the original VPD run intact to maintain faithfulness to the original implementation. In this section, we want to evaluate VPD methods with and without the sparsity adaptive loss to claim that adding sparsity loss would not make the original VPD scalable on large models. The results are in Table \ref{tab:interp_sparse}, \ref{tab:edit_sparse}, \ref{tab:localize_sparse}. We found that adding or removing the sparsity loss does not significantly outperform each other, and both still perform poorly on Gemma-2-2b and Qwen-3-8b.

\begin{table}[h]
    \scriptsize
    \centering
    \caption{Interpretability and Diversity experiment results comparing VPD with and without adaptive sparisty loss.}
    \begin{tabular}{l|c|c|c|c|c|c}
        \toprule
        & \multicolumn{2}{c|}{GPT2} & \multicolumn{2}{c|}{Gemma-2-2b} & \multicolumn{2}{c}{Qwen-3-8b} \\
        Method & Interp $\uparrow$ & Sim $\downarrow$ & Interp $\uparrow$ & Sim $\downarrow$ & Interp $\uparrow$ & Sim $\downarrow$ \\
        \midrule
        VPD with adaptive $L_0$ & 0.34 $\pm$ 0.03 & 0.00 $\pm$ 0.00 & 0.21 $\pm$ 0.02 & 0.24 $\pm$ 0.00 & 0.20 $\pm$ 0.02 & 0.34 $\pm$ 0.01 \\
        VPD without adaptive $L_0$  & 0.37 $\pm$ 0.03 & 0.01 $\pm$ 0.00 & 0.20 $\pm$ 0.02 & 0.44 $\pm$ 0.01 & 0.22 $\pm$ 0.02 & 0.18 $\pm$ 0.00 \\
        \bottomrule
    \end{tabular}
    \label{tab:interp_sparse}
\end{table}

\begin{table}[h]
    \scriptsize
    \centering
    \caption{Weight Editing Localization experiment results comparing VPD with and without adaptive sparisty loss.}
    \begin{tabular}{l|c|c|c|c|c|c}
        \toprule
        & \multicolumn{2}{c|}{GPT2} & \multicolumn{2}{c|}{Gemma-2-2b} & \multicolumn{2}{c}{Qwen-3-8b} \\
        Method & $ratio$ $\uparrow$ & $localization$ $\uparrow$ & $ratio$ $\uparrow$ & $localization$ $\uparrow$ & $ratio$ $\uparrow$ & $localization$ $\uparrow$ \\
        \midrule
        \multicolumn{7}{l}{\textbf{Single}} \\
        \midrule
        VPD with adaptive $L_0$ & 0.2 $\pm$ 0.1 & 0.004 $\pm$ 0.002 & 0.6 $\pm$ 0.1 & 0.005 $\pm$ 0.001 & 0.9 $\pm$ 0.2 & 0.020 $\pm$ 0.003 \\
        VPD without adaptive $L_0$  & 0.9 $\pm$ 0.2 & 0.020 $\pm$ 0.002 & 0.5 $\pm$ 0.1 & 0.004 $\pm$ 0.001 & 0.9 $\pm$ 0.2 & 0.020 $\pm$ 0.003 \\
        \midrule
        \multicolumn{7}{l}{\textbf{Multiple}} \\
        \midrule
        VPD with adaptive $L_0$ & 0.2 $\pm$ 0.0 & 0.003 $\pm$ 0.000 & 0.4 $\pm$ 0.0 & 0.004 $\pm$ 0.000 & 0.8 $\pm$ 0.0 & 0.017 $\pm$ 0.001 \\
        VPD without adaptive $L_0$  & 0.9 $\pm$ 0.1 & 0.015 $\pm$ 0.001 & 0.4 $\pm$ 0.0 & 0.004 $\pm$ 0.000 & 0.8 $\pm$ 0.0 & 0.017 $\pm$ 0.001 \\
        \bottomrule
    \end{tabular}
    \label{tab:edit_sparse}
\end{table}

\begin{table}[h]
    \centering
    \footnotesize
    \caption{Meaning localization experiment results comparing VPD with and without adaptive sparisty loss.}
    \begin{tabular}{l|c|c|c}
        \toprule
        & \multicolumn{1}{c|}{GPT2} & \multicolumn{1}{c|}{Gemma-2-2B} & \multicolumn{1}{c}{Qwen-3-8b} \\
        Method & Matching $\uparrow$ & Matching $\uparrow$ & Matching $\uparrow$ \\
        \midrule
        VPD with adaptive $L_0$ & 0.18 $\pm$ 0.10 & -0.02 $\pm$ 0.12  & 0.05 $\pm$ 0.14 \\
        VPD without adaptive $L_0$  & 0.31 $\pm$ 0.11 & -0.02 $\pm$ 0.10 & 0.17 $\pm$ 0.11 \\
        \bottomrule
    \end{tabular}
    \label{tab:localize_sparse}
\end{table}




\section{IOI Circuit Reverse Engineering}
\label{sec:ioi_circuit_full}
In this section, we outline the full story of how we interpret the IOI circuit \citep{ioi}. We use the original prompt ``When Mary and John went to the store, John gave a drink to".

\subsection{Notation}
We denote $H8.6$ as the 6th attention head at layer 8 of GPT2s.
For a component $c$ of a matrix $W$ with input activation $x_t$ at token $t$, write
$a_c(x_t)=v_c^\top x_t$ for its activation, $g_{t,c}$ for its causal importance gate, and $e_c(x_t)=g_{t,c}\,a_c(x_t)$
for its \emph{effective} activation, so that the module's output is
$\hat y_t=\sum_c e_c(x_t)\,u_c+b$. For a matrix whose output (resp.\ input) is the
concatenation of $H$ head blocks of width $d_h$, we write $u^{h}_{c}\in\R^{d_h}$
(resp.\ $v^{h}_{c}$) for the block of $u_c$ (resp.\ $v_c$) belonging to head $h$.

\textbf{Attribution patching.} We compute the attribution patching to identify the important component as follows. Let $\textsc{Logitdiff} = Logit(Mary) - logit(John)$ be the logit difference between the two candidate answers. To every
component $c$, we attach a multiplier $\xi_{t,c}$ at token $t$ to compute gradient:
\[
  \tilde y_t \;=\; y_t \;+\; \sum_c (\xi_{t,c}-1)\,e_c(x_t)\,u_c ,
\]
holding $e_c(x_t)$ frozen. At $\xi\equiv 1$ the added term is zero, and hence the forward pass is unchanged and $\tilde y_t = y_t$ exactly. The attribution score of component $c$ at token $t$ is then
\[
  attrib(c,t)\;=\;\left.\frac{\partial \textsc{Logitdiff}}{\partial \xi_{t,c}}\right|_{\xi\equiv 1},
\]
obtained from computing one forward and one backward pass. We rank the component per weight matrix and by $|attrib(c,t)|$.

\textbf{QK contribution.} We compute the contribution of a component $c_1$ at $Q$ matrix on token $t_1$ and $c_2$ at
$K$ matrix on token $t_2$ ($t_2\le t_1$), within head $h$, as
\[
\mathrm{contrib}^{QK}_{h}(c_1,t_1;\,c_2,t_2)\;=\;\frac{1}{\sqrt{d_h}}\, 
\Big|e_{c_1}(x_{t_1})\,e_{c_2}(x_{t_2})\,\big\langle u^{h}_{c_1},\,u^{h}_{c_2}\big\rangle\Big| ,
\]
assuming that both components fired ($g_{t_1,c_1} > 0$ and $g_{t_2,c_2}>0$), so that
$e_c=a_c$; the higher the contribution the more important the component pair to the $QK$
circuit.

\textbf{OV contribution.} Furthermore, for $OV$ circuits, we compute the contribution of a component $c_1$ at $V$ matrix on token $t_1$ to a component $c_2$ at $O$ matrix on token $t_2$ of the same attention head as:
\[
\mathrm{contrib}^{OV}_{h}(c_1\!\to\!c_2)[t_1,t_2]\;=\;
pattern^{h}[t_1,t_2] \cdot
 \; e_{c_1}(x_{t_1})\cdot
\big\langle u^{h}_{c_1},\,v^{h}_{c_2}\big\rangle \cdot \;g_{t_2,c_2}
\]

\textbf{Cross layer contribution.} Similarly, for matrix $O$ at lower layer and matrix $Q,K,V$ of higher layer attention heads, we define the
contribution as:
\[
\mathrm{contrib}^{\mathrm{res}}(c_1\!\to c_2) = 
e_{c_1}(x_t)\cdot\big\langle u_{c_1},\,v_{c_2}\big\rangle\cdot g_{t_2,c_2},
\]
for $c_1$ a component of $O$ at layer $\ell$ and $c_2$ a component of $Q$, $K$ or $V$
at layer $\ell'>\ell$.

\textbf{QK weight editing.} Lastly, we define an \emph{edit} by deleting components from a parameter. For $\mathcal E_Q$ a set of components of
$Q$ and $\mathcal E_K$ a set of components of $K$, we set
\[
W^{Q}_{edit}=W^{Q}-\!\!\sum_{c\in\mathcal E_Q}\!\!P^{Q}_{c},
\qquad
W^{K}_{edit}=W^{K}-\!\!\sum_{c\in\mathcal E_K}\!\!P^{K}_{c},
\]
and recompute the pattern of head $h$ from the
edited matrices,
\[
pattern^{h}_{edit}[t_1,t_2]\;=\;\mathrm{softmax}_{\,t_2\le t_1}
\left(\frac{1}{\sqrt{d_h}}\,
\big(W^{Q}_{edit}x_{t_1}\big)^{h\top}\big(W^{K}_{edit}x_{t_2}\big)^{h}\right).
\]
We then report
$\Delta pattern(edit) = pattern - pattern_{edit}$, which represents how the pattern
changes under the edit.

\paragraph{Other notes:}
(1) A LayerNorm lies between the residual write at
layer $\ell$ and the read at layer $\ell'$, however, for simplicity, we report the \emph{un-normalised}
composition $\langle u_{c_1},v_{c_2}\rangle$ rather than folding that LayerNorm into the
read direction. 
(2) We use ASPD on GPT2s with a shared causal importance function for $Q,K,V$ matrices trained at residual-stream-pre (right before the input of the attention head), and for $O, MLP_{in}, MLP_{out}$ at the residual-stream-mid (right before the input of the MLP); this causes the same components with the index at the $Q, K, V$ or $O, MLP_{in}, MLP_{out}$ of the same layer to fire on the same token set and therefore mean the same concept. Other training details of this experiment are in Appendix \ref{sec:gpt2_decomp}.

\subsection{Story}

\textbf{Background.} In the paper \citet{ioi}, researchers show one of the first attempts to reverse engineer by localizing what each module (attention head) does in a prompt ``When Mary and John went to the store, John gave a drink to", in which the GPT2s model correctly predicts the next token is `` Mary". They found that some attention heads are strongly important for this task, and they classified the attention heads into 6 main class ``Previous token heads, Duplicate token heads, Induction heads, S-inhibition heads, Name mover heads, Negative name mover heads" and an addition class ``Backup name mover heads" only appears when ablating the class ``Name mover heads". \citet{ioi} provided a blueprint for the IOI circuit at the attention head level but never what the underlying computations mean.

\textbf{Previous Token Heads.} We started with the blueprint of the IOI circuit (see Figure \ref{fig:ioi_circuit_mech} or Figure 2 in \citet{ioi}) and looked at the first important module: ``Previous token heads" which contains $H2.2, \, H4.11$. We chose $H4.11$ for illustration as it shows the strongest attention pattern of ``Previous token heads" behavior; however, we can interpret $H2.2$ the same way. \textbf{QK.} Based on the blueprint, the $H4.11$ always attends to the token right before, and the computation of the head at the position $Q:4(went)-K:3(John)$ is the most important for the circuit to predict the output. Therefore, in Figure \ref{fig:qk_h4_11}, we looked at which components in the $QK$ matrix most strongly reconstruct the attention pattern at position $Q:4(went)-K:3(John)$. We found that the dense components (components $12Q, 12K$ and $70Q, 70K$ with fire frequency of 8.5\% and 23.190\% respectively and activate on all tokens on the IOI prompt except for the first token) of the $Q$ and $K$ matrices interact strongly with each other at the position, contributing the top 4 strongest pairs. Furthermore, non-dense component interactions of $5740Q, 440Q$ (``follow a name") and $341K$ (``names") are also the strongest pairs that contribute to the attention pattern at the position. We edited the dense components and non-dense components in Figure \ref{fig:qk_h4_11} and found that attention patterns change significantly, suppressing attention to the previous tokens at many positions. Notably, editing the non-dense components leads to the most significant change in the attention pattern of $Q:4(went)-K:3(John)$, suggesting that the model recognizes that the token $4(went)$ follows after a name. \textbf{OV.} In the blueprint, $H4.11 \;OV$ moves the information from token $3(John)$ to token $4(went)$. As illustrated in Figure \ref{fig:ov_h4_11}, we ran attribution patching and found two interesting components: $2774V$ (``John", top-1) and $341V$ (``names", top-8) fire on $3(John)$. Those components contribute the most to the components $5608O$ (``follow names like ``John", ``James", ``Joan", etc.") and $5690$ (``follow a name") respectively on $4(went)$, signaling that the head moves the information of ``previous token is name" and ``previous name is John" to the residual stream for downstream layers.

\textbf{Induction heads.} Which takes the output of $H4.11$ at token $4(went)$? $H5.5$. \textbf{Input.} In Figure \ref{fig:v_in_h5_5}, the component $5897V\,H5.5$ (``token followed a name") at position $4(went)$ is among the top-5 components that the component $5690O\,H4.11$ (``token followed a name") contributes most strongly to. \textbf{QK.} We analyzed the attention pattern of $H5.5$ at position $Q:9(John)-K:4(went)$ that exhibits the induction behavior in Figure \ref{fig:qk_h5_5}. We found that $164Q$ (``names") and $5897K$ (``token followed a name") contribute the most to the attention position, and editing them removed the induction pattern entirely. This suggests that the head reads the name and the token following a name to produce the induction pattern. \textit{Limitation:} our method did not learn the component that specifically represents the ``token after John" signal, making our explanation here less localized compared to what we would want; increasing the number of components per weight matrix may surface more fine-grained components. \textbf{OV.} See Figure \ref{fig:ov_h5_5}. We ran attribution patching at position $9(John)$ and found component $228O$ that we suspected it is induction component based on its activation. To test the hypothesis, we ran an induction probe (detailed in Appendix \ref{sec:probe-induction}) and found that it indeed strongly exhibits induction behavior, in which the result shows that it is the top-5 $probe\_score$ that has $mass_{5,288} \geq 0.25$. We also found components $729O, 665O, 350O$ that are induction components (with probe scores of top-2, top-6, and top-7, respectively, all fire on the IOI prompt and $mass \geq 0.25$) but with lower attribution patching scores. Inspecting those components, we found that components $228O, 729O, 665O$ are strongly influenced by $5890V$ (``token followed a name"), which is also the component strongly fired by $5690O\,H4.11$. This shows the mechanism of connection between ``Previous token heads" and ``Induction head" in the model and the mechanism of writing of induction signal to the activation space in $H5.5$.

\begin{figure}[h]
    \centering
    \includegraphics[width=0.9\linewidth]{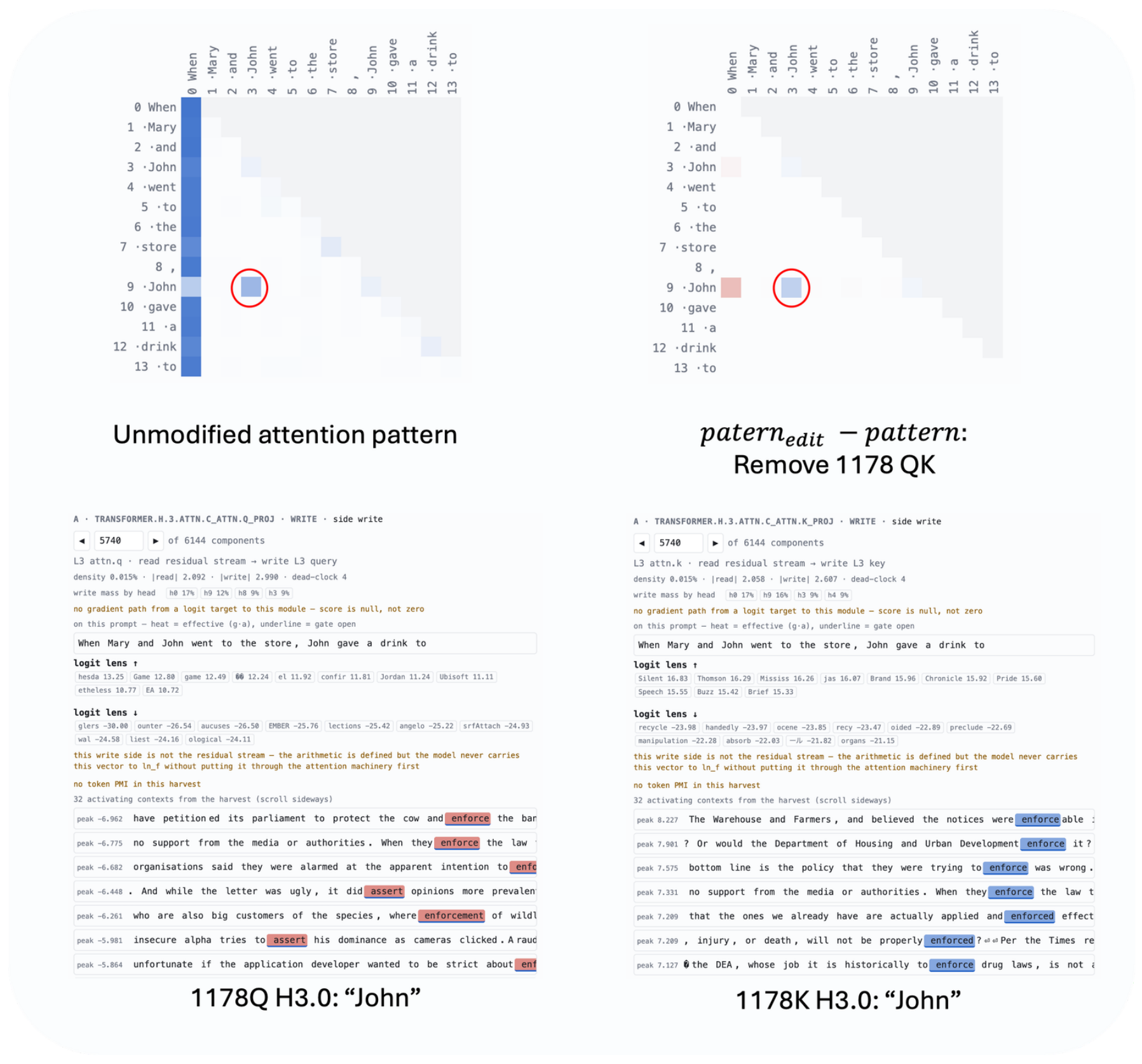}
    \caption{$QK$ circuit at position $Q:9(John)-K:3(John)$ of $H3.0$ (``duplicate token head"). We observed that components $1178QK$ (``John") contribute strongly to the attention pattern, and ablating those components suppresses the ``duplicate token" attention pattern.}
    \label{fig:qk_h3_0}
\end{figure}

\begin{figure}[h]
    \centering
    \includegraphics[width=0.9\linewidth]{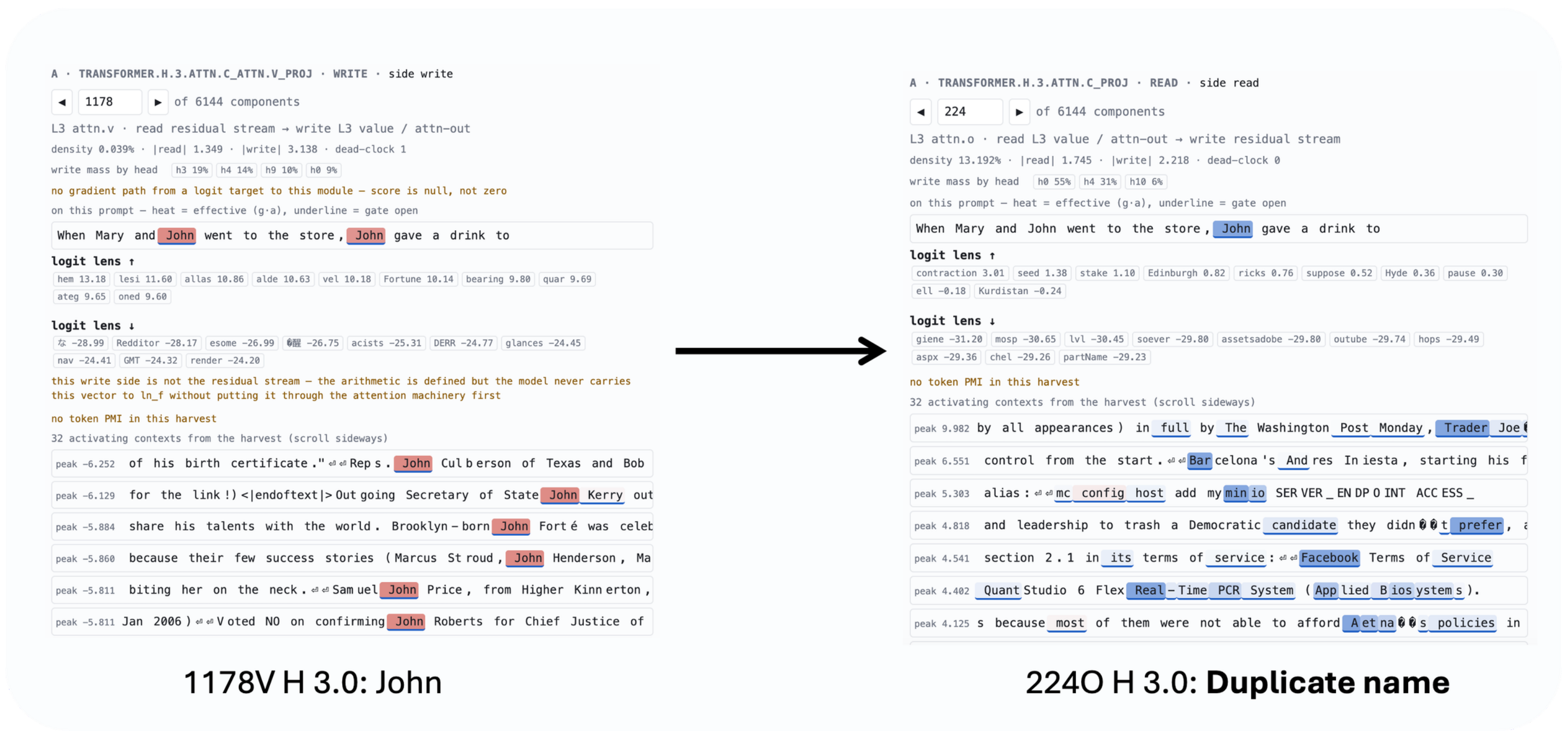}
    \caption{$OV$ circuit of $H3.0$ (``duplicate token head"). The components $1178V$ (``John") and $224O$ (``duplicate name") have a strong causal effect on each other. The component $224O$ writes the ``duplicate token" signal to the activation space, which is the main role of $H3.0$.}
    \label{fig:ov_h3_0}
\end{figure}

\begin{figure}[h]
    \centering
    \includegraphics[width=0.9\linewidth]{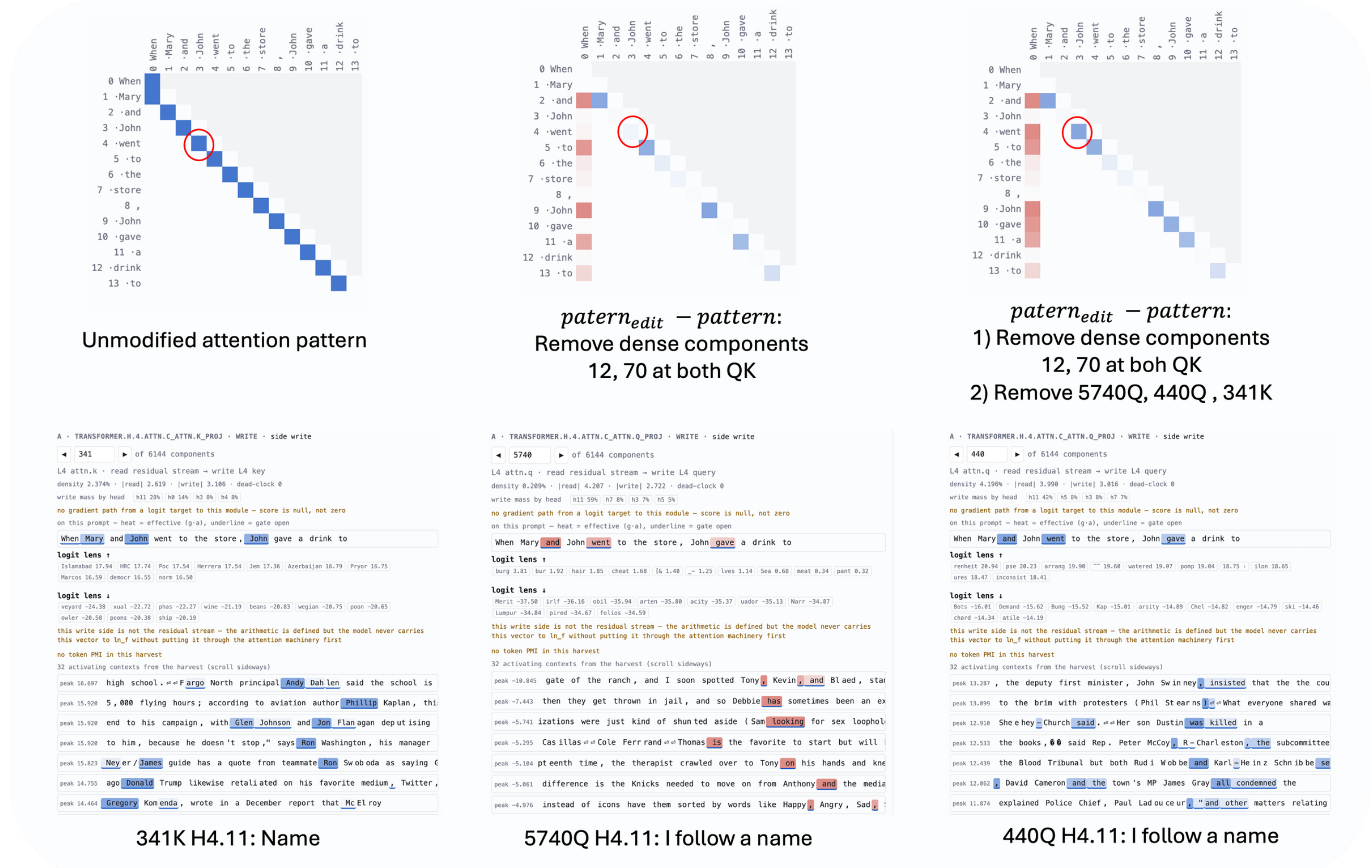}
    \caption{$QK$ circuit at position $Q:4(went)-K:3(John)$ of $H4.11$ (``previous token head"). We observed that components $12QK,70QK$ (dense components), $5740Q, 440Q$ (``follow a name"), $341K$ (``name") contribute strongly to the attention pattern, and ablating those components suppresses the ``previous token" attention pattern.}
    \label{fig:qk_h4_11}
\end{figure}

\begin{figure}[h]
    \centering
    \includegraphics[width=0.8\linewidth]{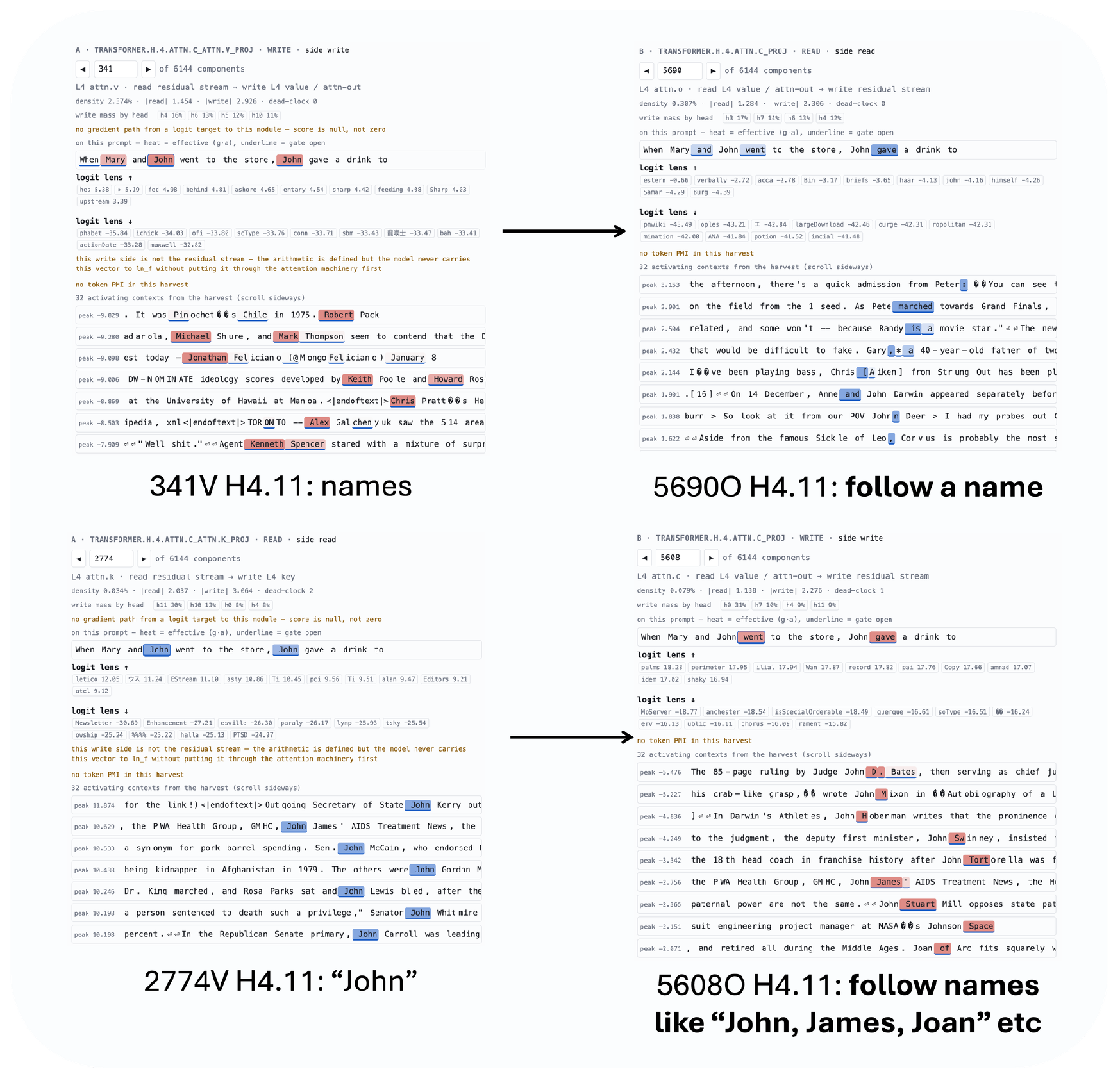}
    \caption{$OV$ circuit of $H4.11$ (``previous token head"). The components $341V$ (``names"), $2774V$ (``John") have a strong causal effect on $5690O$ (``follow a name"), $5608O$ (``follow names like John, James, Joan, etc."). The components $5690O, 5608O$ write the ``follow a name (John)" signal to the activation space, which is the main role of $H4.11$.}
    \label{fig:ov_h4_11}
\end{figure}

\begin{figure}[h]
    \centering
    \includegraphics[width=0.9\linewidth]{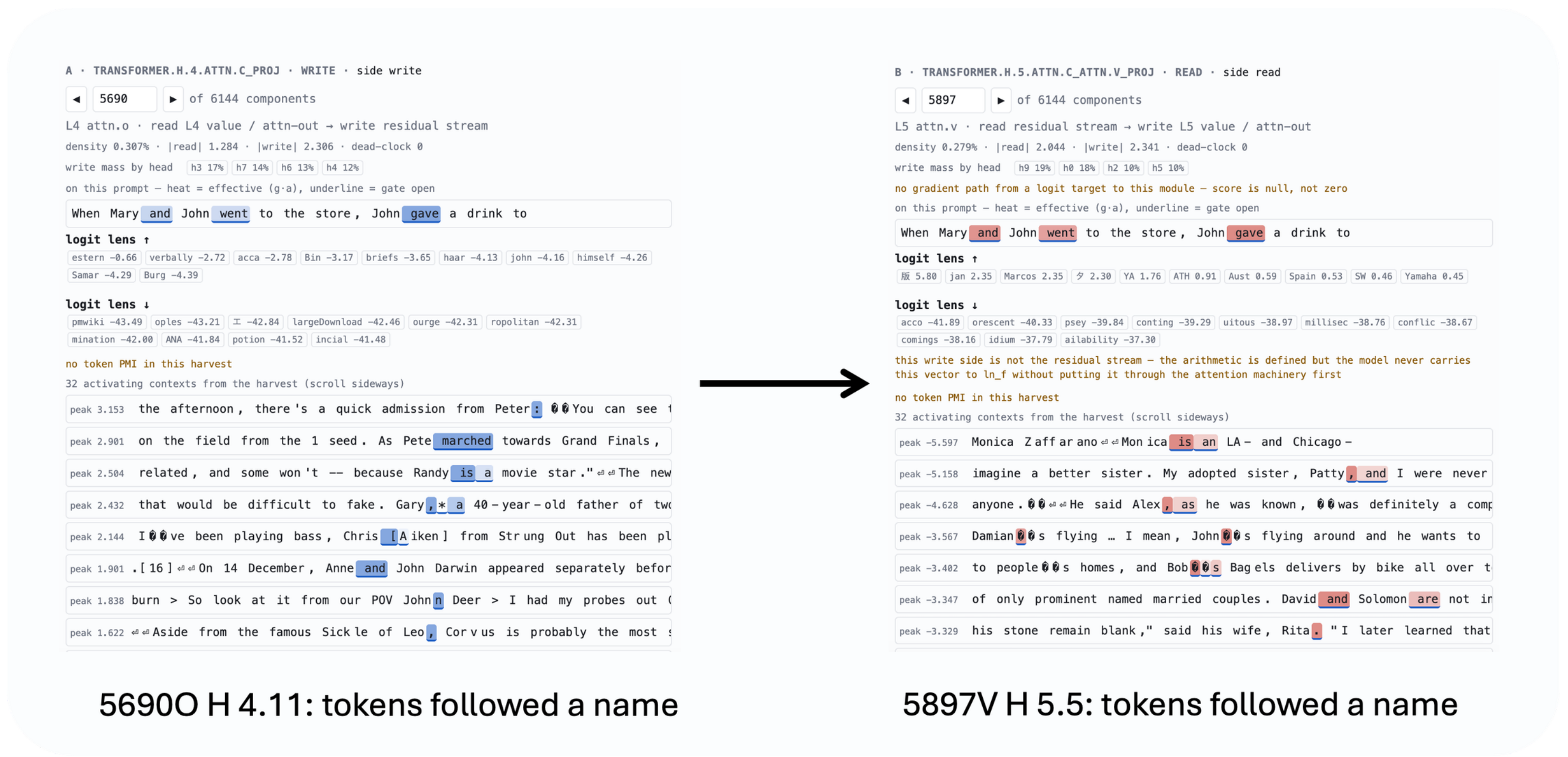}
    \caption{Connection between $H4.11$ (``previous token head") and $5.5$ (``induction head"). Component $5690O\, H4.11$ (``tokens followed a name"), which is the component that writes the main functionality of $H4.11$ to the activation space, contributes strongly to the component $5897V \, H5.5$ (``tokens followed a name").}
    \label{fig:v_in_h5_5}
\end{figure}

\begin{figure}[h]
    \centering
    \includegraphics[width=0.9\linewidth]{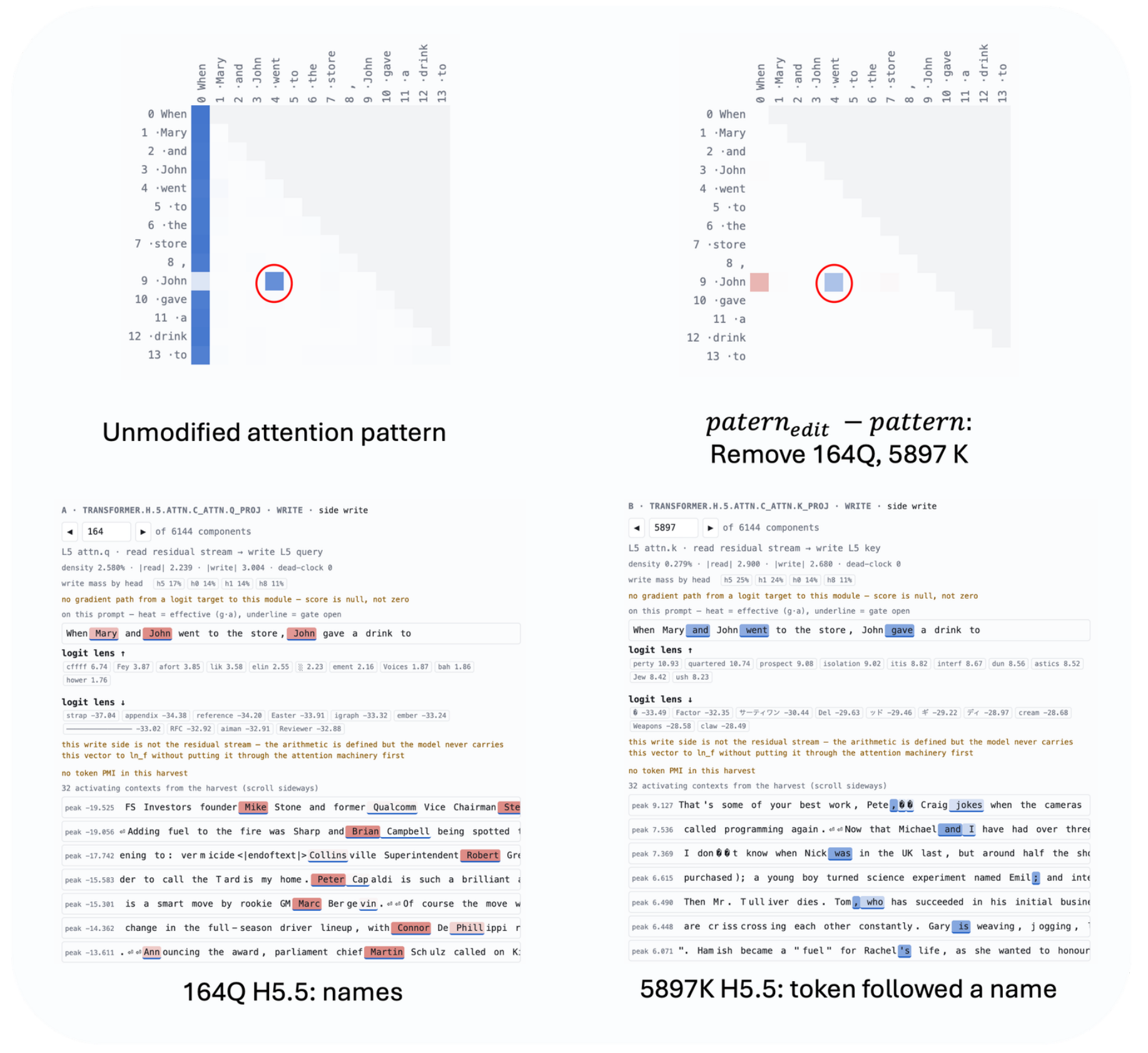}
    \caption{$QK$ circuit at position $Q:9(John)-K:4(went)$ of $H5.5$ (``induction head"). We observed that components $164Q$ (``names"), $5897K$ (``token followed a name") contribute strongly to the attention pattern and ablating those components suppress the ``induction" attention pattern.}
    \label{fig:qk_h5_5}
\end{figure}

\begin{figure}[h]
    \centering
    \includegraphics[width=\linewidth]{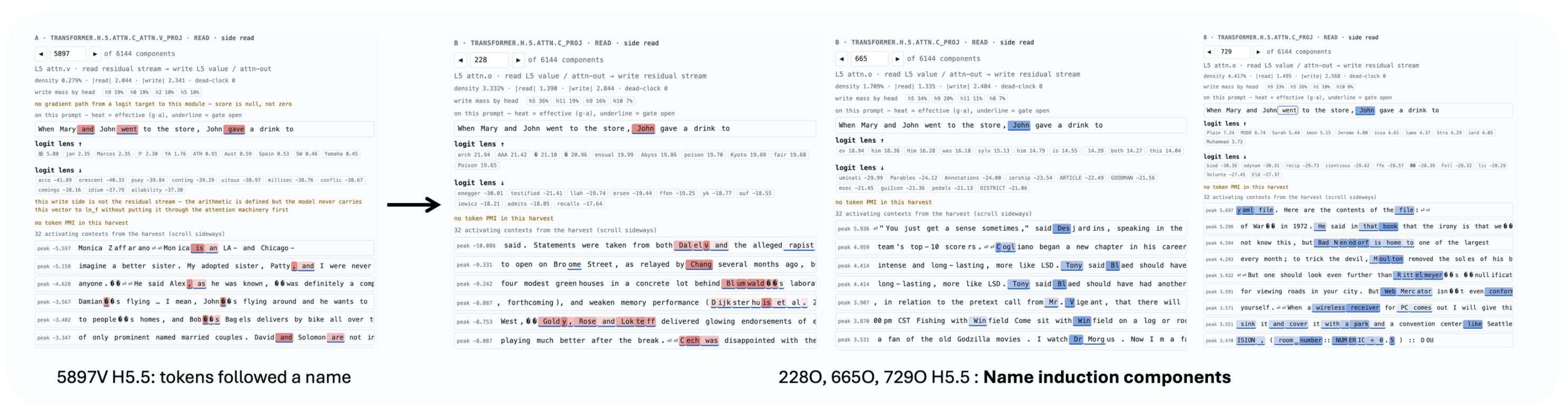}
    \caption{$OV$ circuit of $H5.5$ (``induction head"). The components $4897V$ (``tokens followed a name") have a strong causal effect on $228O, 665O, 7290O$ (``name induction"). The components $228O, 665O, 7290O$ write the ``induction" signal to the activation space, which is the main role of $H5.5$.}
    \label{fig:ov_h5_5}
\end{figure}

\textbf{Duplicate token heads.} Another important class is ``Duplicate token heads"; we inspect the module $H3.0$ that most strongly exhibits this behavior. \textbf{QK.} In Figure \ref{fig:qk_h3_0}, we inspected the attention pattern on the repeated name $Q:9(John)-K:3(John)$ and saw that the component $1178Q, V$ mainly fire on ``John" tokens contribute the most to the attention pattern, ablating them suppressed the original attention patterns significantly. This means that the head recognizes the repeated ``John" name and attend to the previous position. \textbf{OV.} See Figure \ref{fig:ov_h3_0}. We ran attribution patching and found $224O$ with the top-5 attribution score, and it seems to exhibit ``duplicate token" behavior where it fires on the repeat of names or tokens. We ran the test on the probe in Appendix \ref{sec:probe-duplicate} and observed that $224O$ indeed is the top-1 and top-2 in duplicate token and duplicate name probe scores, respectively, while having $mass_{0,224} \geq 0.25$. We also found that the component $1178V$ (``John") is the top-2 contributor to the activation of component $224O$. All of these show the computation of $H3.0$ in recognizing the repeated token and output the ``duplicate token" signal to the activation space.

\begin{figure}[h]
    \centering
    \includegraphics[width=0.9\linewidth]{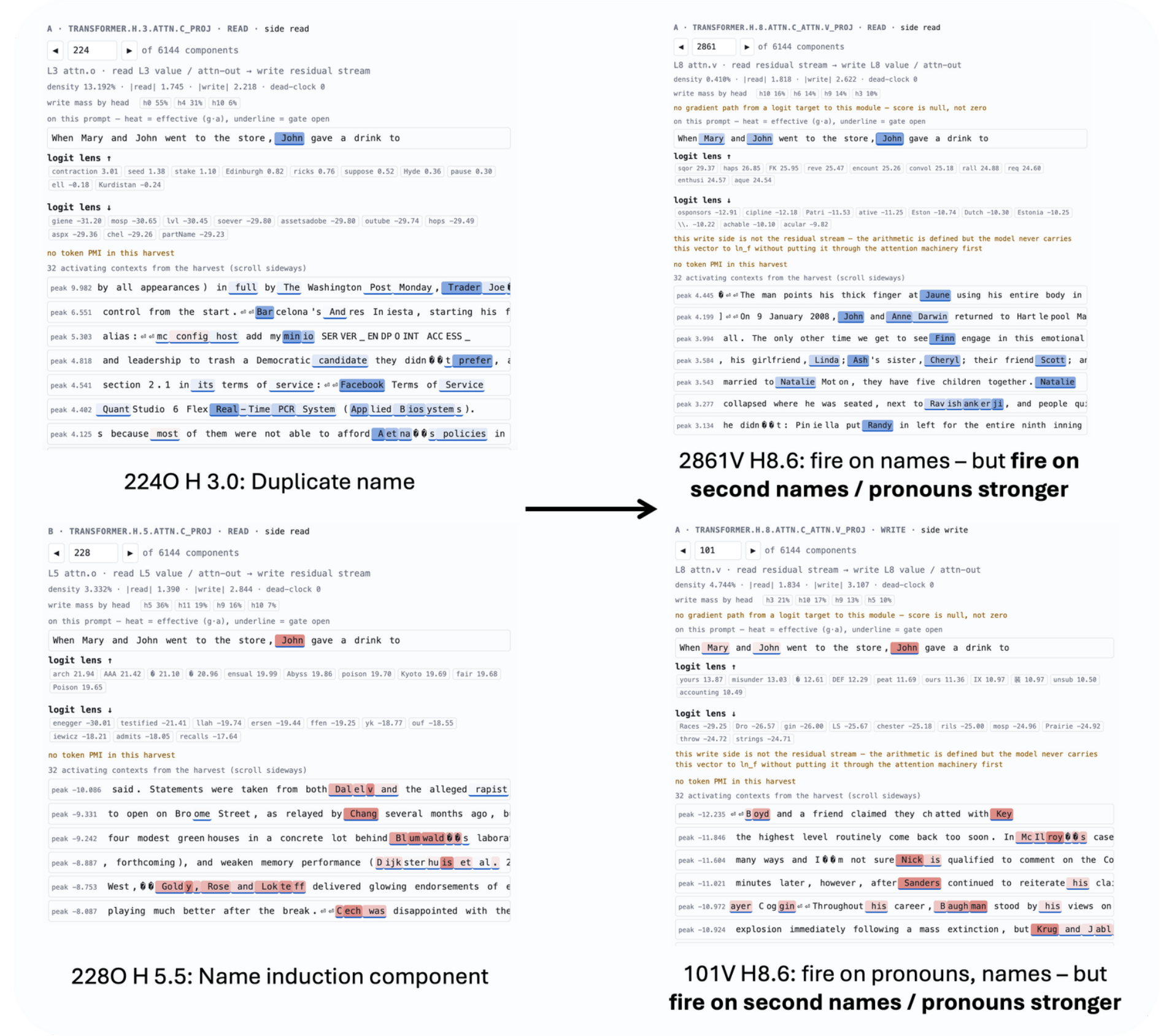}
    \caption{Connection between $H3.0$ (``duplicate token head"), $5.5$ (``induction head") and $H8.6$ (``S-inhibition head"). Components $224O\, H3.0$ (``duplicate name") and $228O\, H5.5$ (``name induction"), which are the components that write the main functionality of $H3.0$ and $H5.5$ to the activation space, contribute strongly to the component $2861V, 101V \, H8.6$ (``fire on pronouns, names - but \textit{fire on second names/pronouns stronger}").}
    \label{fig:v_in_h8_6}
\end{figure}

\begin{figure}[h]
    \centering
    \includegraphics[width=0.9\linewidth]{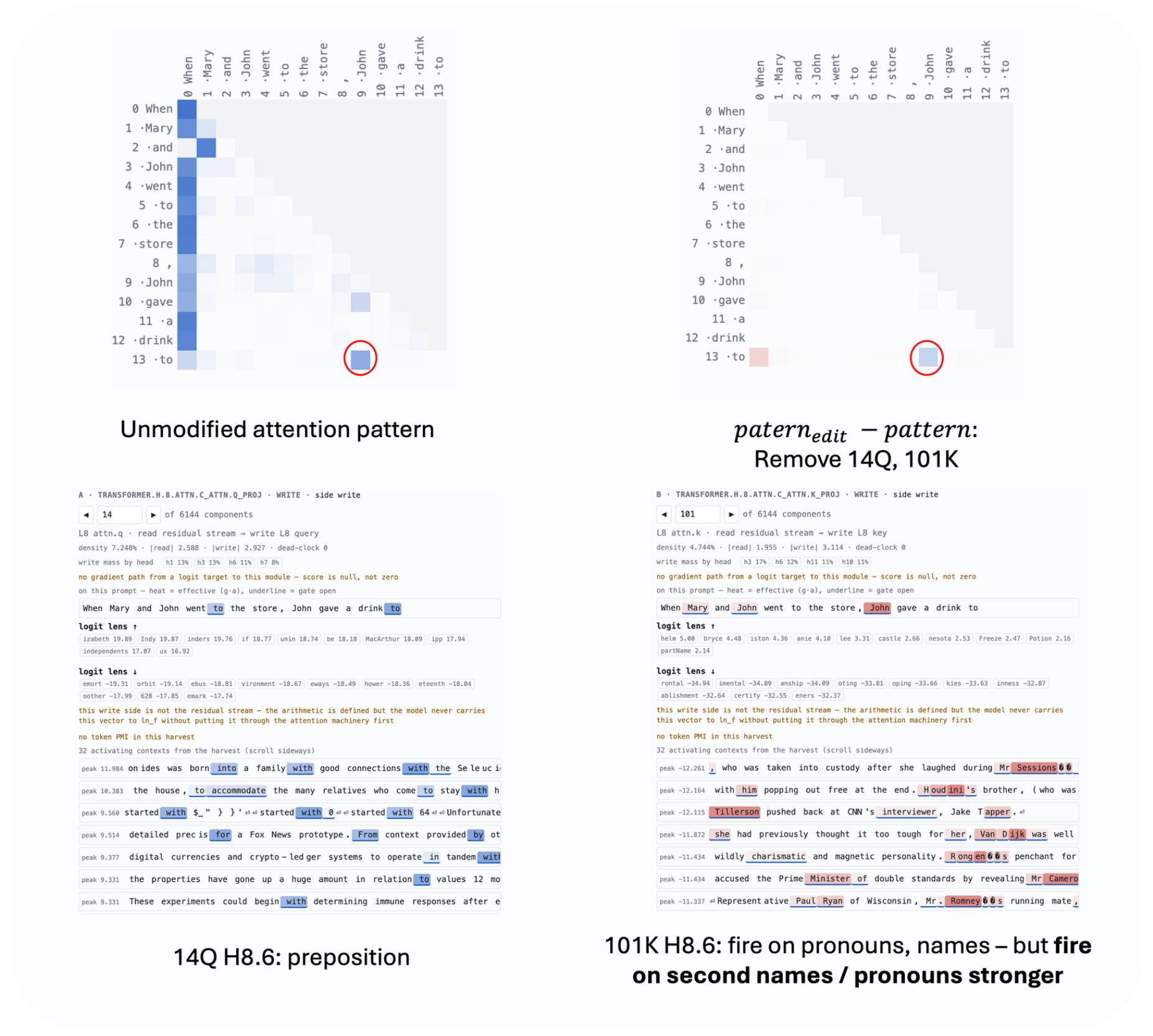}
    \caption{$QK$ circuit at position $Q:13(to)-K:9(John)$ of $H8.6$ (``S-inhibition head"). We observed that components $14Q$ (``preposition"), $101K$ (``fire on pronouns, names - but \textit{fire on second names/pronouns stronger}") contribute strongly to the attention pattern, and ablating those components suppresses the ``S-inhibition" attention pattern.}
    \label{fig:qk_h8_6}
\end{figure}

\begin{figure}[h]
    \centering
    \includegraphics[width=1\linewidth]{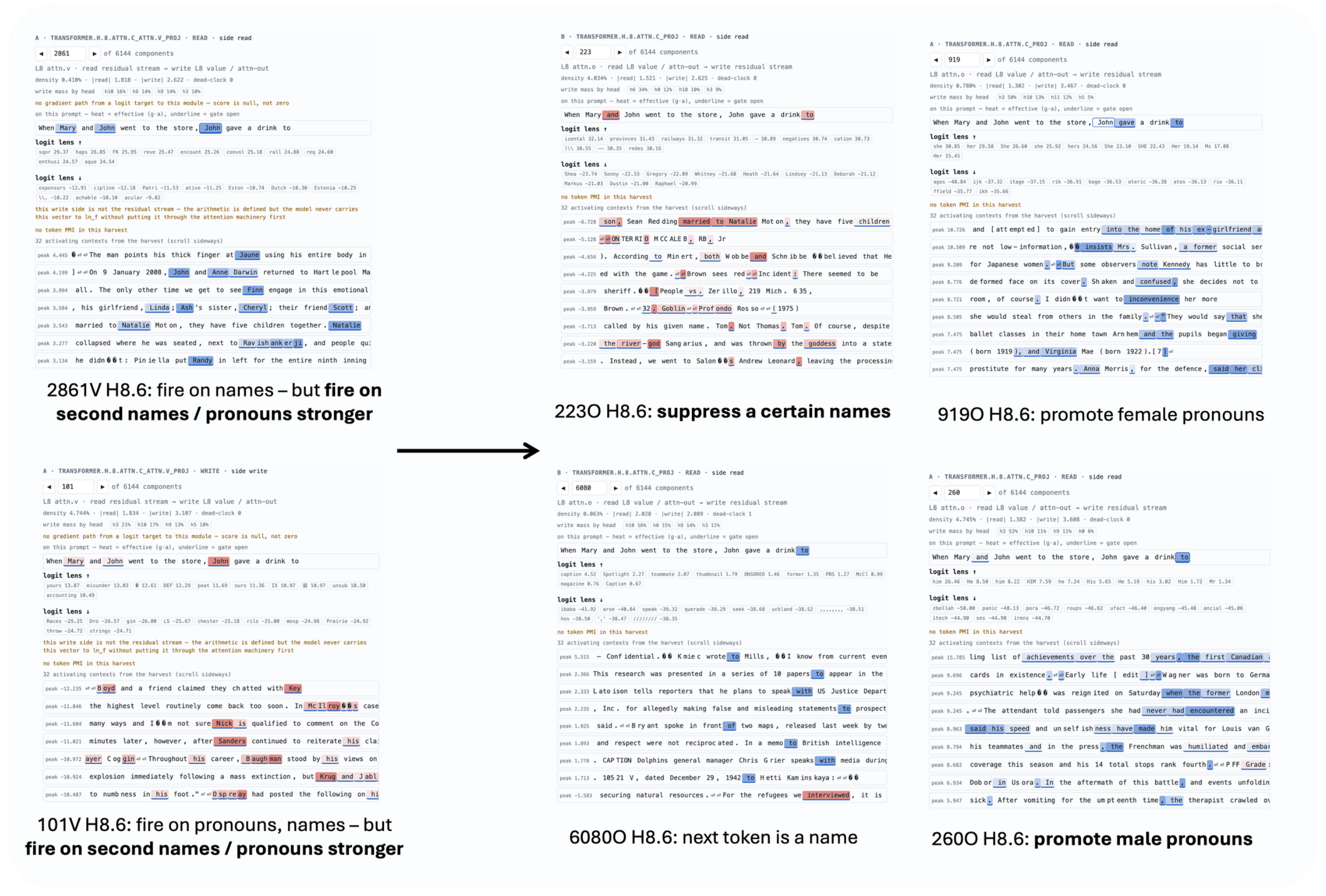}
    \caption{$OV$ circuit of $H8.6$ (``S-inhibition head"). The components $2861V, 101V$ (``fire on pronouns, names - but \textit{fire on second names/pronouns stronger}") have a strong causal effect with $223O$ (``suppress certain names"), $919O$ (``promote female pronouns"), $6080O$ (``next token is a name"), $260O$ (``promote male pronouns"). The components $223O, 919O, 6080O, 260O$ suppress/promote ``male/female pronouns '' and write the information to the activation space, which is the main role of $H8.6$.}
    \label{fig:ov_h8_6}
\end{figure}

\textbf{S-inhibition heads.} Which head takes the output of $H5.5, H3.0$? $H8.6$. \textbf{Input.} In Figure \ref{fig:v_in_h8_6}, we observed that 2 components that are important for OV circuit of $H8.6$, $2861V$ and $101V$ at position $9(John)$, strongly activated by the components $224O\, H3.0$ (``Duplicate name", top-3 contribution to $2861V\,H8.6$ and top-4 to $101V\,H8.6$ among all $H3.0$ components) and $228O\,H5.5$ (``Name induction", top-3 contribution to $2861V\,H8.6$ and top-1 to $101V\,H8.6$ among all $H5.5$ components). This suggests that $H8.6$ processes the information of duplicated names and induction from upstream heads. \textbf{QK.} Similarly to before, in Figure \ref{fig:qk_h8_6}, we investigated the attention pattern at $Q:13(to)-K:9(John)$, found that the pair $14Q$ (``preposition") and $101K$ (``fire on pronouns, names – but fire on second names/pronouns stronger") have a strong contribution, and ablating them changed the attention pattern at the target position significantly. This demonstrates that $H8.6$ recognizes the repeated name and attends to it. \textbf{OV.} In figure \ref{fig:ov_h8_6}, we ran attribution patching at position $13(to)$ and found 3 interesting components that are among the highest scores: $919O$ (``promote female pronouns", top-1), $260O$ (``promote male pronouns", top-6), and $6080O$ (``next token is a name", top-10). We traced the components in the $V$ matrix of $H8.6$ at position $9(John)$ and saw $2861V, 101V$ (``fire on names - but fire on second names/pronouns stronger") consistently among the strongest contributors to the 3 $O$ matrix components. We also saw that the $101V$ promotes component $223O$, which suppresses certain names and will be the input of ``Name mover heads, Negative name mover heads". Additionally, we tested our hypothesis of ``activate on second names/pronouns stronger" by running a probe in Appendix \ref{sec:probe-factorial}, and found that $101V$ has a high score on both prompt A and prompt B while $2861V$ has a high score on prompt B, confirming our hypothesis. All of this implies that the ``S-inhibition" head recognizes the induction/repeated name signal from previous layers and suppresses or promotes certain pronouns.

\textbf{Name mover heads.} We investigate $H9.6$; other heads can be interpreted similarly. \textbf{Input.} \citet{ioi} shows that ``S-inhibition heads" are input to the Query matrix of ``Name mover heads". Hence, we investigated the important components of the $QK$ circuit of $H9.6$: $122Q$ and $254Q$. We found that both of these components are activated by $223O\,H8.6$, $260O\,H8.6$, $6080O\,H8.6$ among all components of $H8.6$. This confirms the mechanism of $H8.6$ to write name/pronoun suppression to the attention pattern of $H9.6$. Importantly, the important components in the $QK$ circuit of $H9.6$ do not take $919O\,H8.6$ (``promote female pronouns") as input (low contribution compared to other components), which again confirms the mechanism of suppressing attention to $John$ of $H8.6$; however, we did not investigate deeply why $919O\,H8.6$ is not used; a possible direction would be to investigate the ABC prompt \citep{ioi}. \textbf{QK.} In the blueprint, ``Name mover heads" move the name $1(Mary)$ to the last token to predict. We therefore investigate at position $Q:13(to)-K:1(Mary)$ and found that, in Figure \ref{fig:qk_h9_6}, the attention pattern is contributed by $254Q$ (``prepositions, verbs connected to a person"), $122Q$ (``contexts around a certain name"), and $293K$ (``name") the most strongly, and ablating those pairs shifts the attention pattern away from $1(Mary)$ (Figure \ref{fig:qk_h9_6}). \textbf{OV.} We ran attribution patching and found that component $740V$ (``female, woman") at $1(Mary)$ with the strongest attribution score. In Figure \ref{fig:ov_h9_6}, we observed that this component contributes the most to: suppressing $17O$ (``predict next token is a male name", top-1) and promoting $3796O$ (``predict next token is a woman/female name", top-4). This demonstrates the mechanism of the $H9.6$, which promotes female and suppresses male pronouns/names. Limitation: we did not find any ``Mary-specific" component in our decomposition. 

\textbf{Negative name mover heads.} Very similar to ``Name mover heads", we observe the opposite in $H11.10$. \textbf{Input.} In Figure \ref{fig:v_in_h11_10}, we also observed that $223O\,H8.6$, $260O\,H8.6$, $6080O\,H8.6$ contributes the most to the $238Q, 48Q, 263Q$ that are important to $QK$ circuit of $H11.10$. We also found that $919O\,H8.6$ are not among the strongest contributors to the $QK$ circuit of $H11.10$, again showing that the ``Negative name mover heads" only use the ``S-inhibition" signal from $H8.6$. \textbf{QK.} The components $238Q, 48Q, 263Q$ (``prepositions, verbs) and $267K$ (``names") at position $Q:13(to)-K:1(Mary)$ contributed the most; we tested the ablation to confirm the attention pattern shift in Figure \ref{fig:qk_h11_10}. \textbf{OV.} We found that $1323V$ (``female pronouns") and $267V$ (``names") are among the highest-scoring components using attribution patching. These components promote and suppress predicting the next name the most strongly (Figure \ref{fig:ov_h11_10}).

\textbf{Limitation.} Although we believe that we have made good progress in explaining the mechanisms of the IOI circuit, a few limitations exist in our story. (1) There are some mechanisms are overly broad while we would want a more localized meaning, such as identifying ``John" or ``Mary" specific interaction in $H5.5$ or $H9.6$. (2) There are likely many components that we did not fully understand in the fire pattern; for example, there are many different ``token followed a name" components, but only a few of them are important. This could be due to feature splitting \citep{monosemanticity, absorption}, or they genuinely have more intricate meaning that we have not yet discovered. (3) We did not explore how the circuit works on the ABC prompt \citep{ioi}; exploring this might give additional insights into the IOI circuit. 

\begin{figure}[h]
    \centering
    \includegraphics[width=0.9\linewidth]{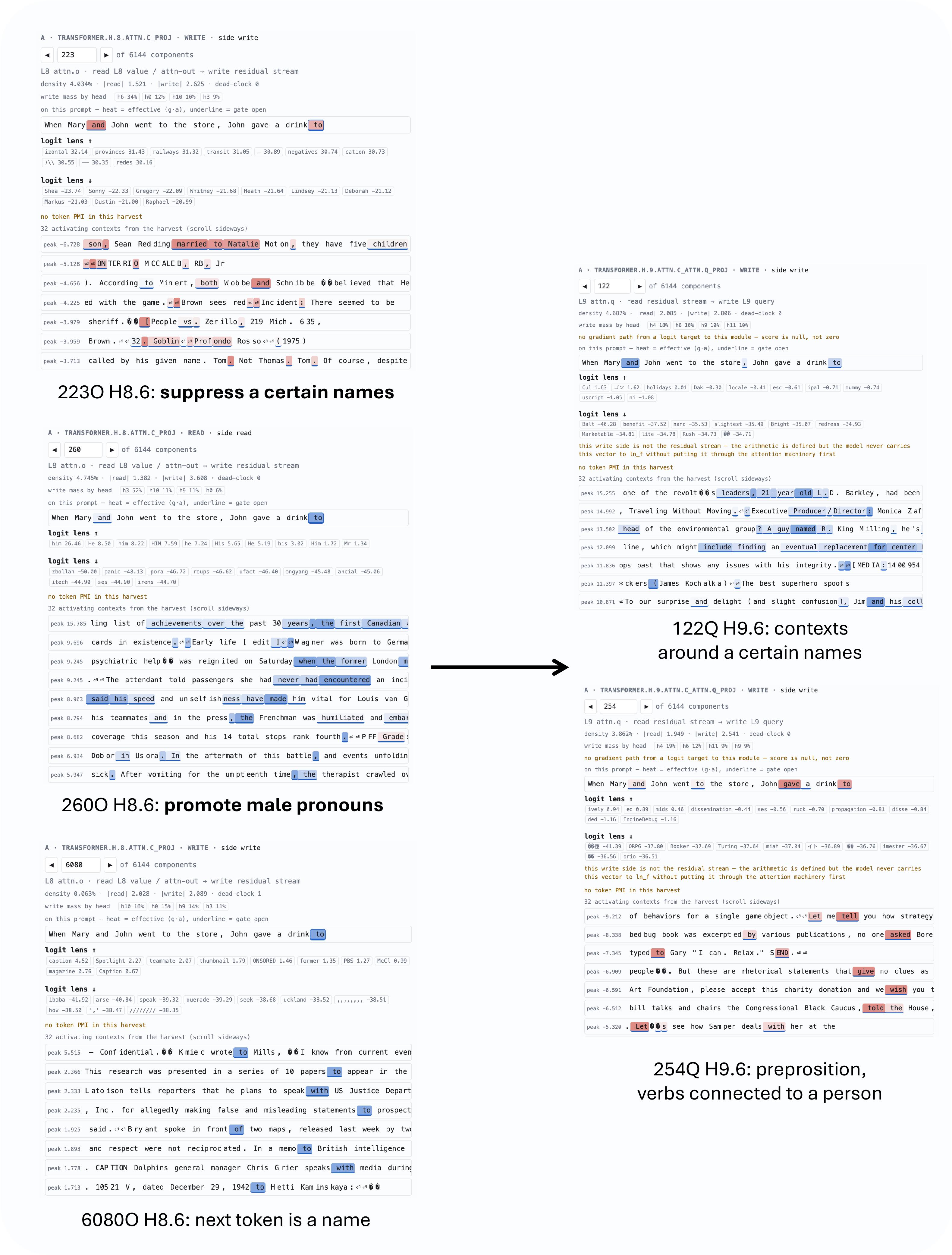}
    \caption{Connection between $H8.6$ (``S-inhibition head") and $H9.6$ (``name mover head"). Components $223O\, H8.6$ (``suppress a certain names"), $260O\, H8.6$ (``promote male pronouns"), $6080O \, H8.6$ (``next token is a name"), which are the components that write the main functionality of $H8.6$ to the activation space, contribute strongly to the component $122Q \, H9.6$ (``contexts around a certain names") and $254Q \, H9.6$ (``preposition, verbs connected to a person").}
    \label{fig:v_in_h9_6}
\end{figure}

\begin{figure}[h]
    \centering
    \includegraphics[width=1\linewidth]{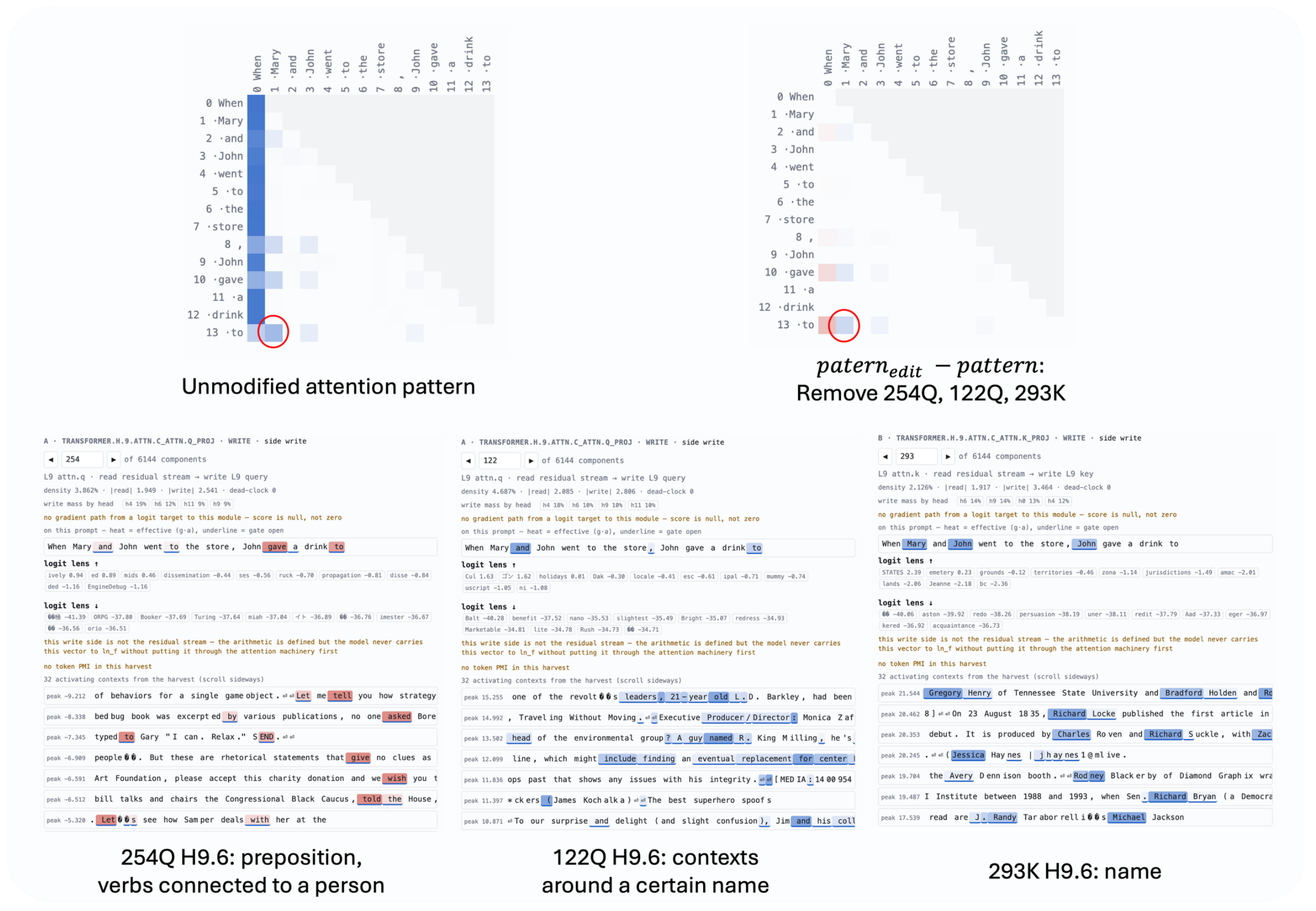}
\caption{$QK$ circuit at position $Q:13(to)-K:1(Mary)$ of $H9.6$ (``name mover head"). We observed that components $254Q$ (``prepositions, verbs connected to a person"), $122Q$ (``contexts around a certain name"), $293K$ (``names") contribute strongly to the attention pattern, and ablating those components suppresses the ``name mover" attention pattern.}
    \label{fig:qk_h9_6}
\end{figure}

\begin{figure}[h]
    \centering
    \includegraphics[width=0.9\linewidth]{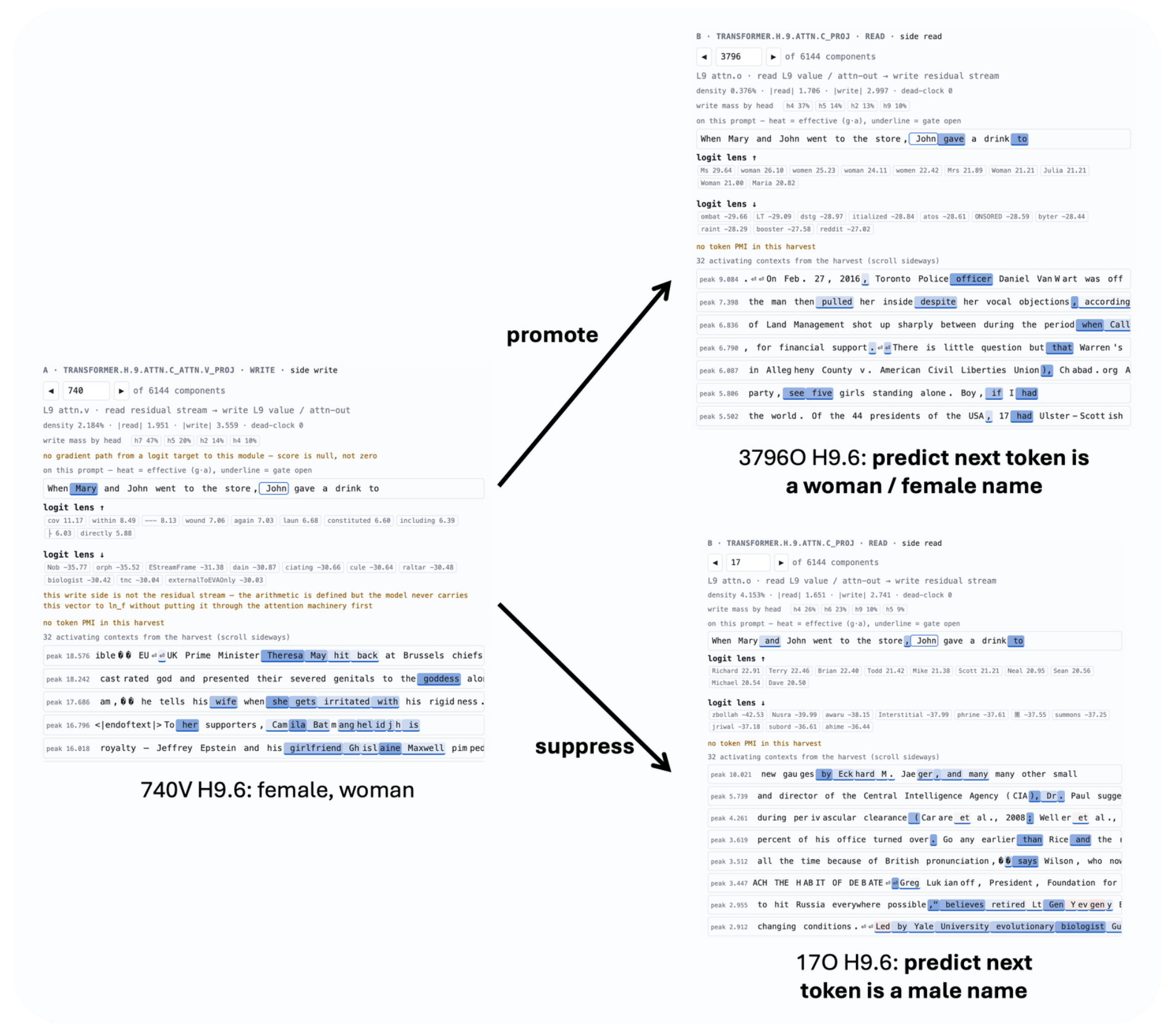}
    \caption{$OV$ circuit of $H9.6$ (``name mover head"). The components $740V$ (``female, woman") have a strong causal effect on $3796O$ (``predict next token is a woman/female name"), $17O$ (``predict next token is a male name"). The components $3796O, 17O$ promote ``male/female pronouns '' and write the information to the activation space, which is the main role of $H9.6$.}
    \label{fig:ov_h9_6}
\end{figure}

\begin{figure}[h]
    \centering
    \includegraphics[width=0.8\linewidth]{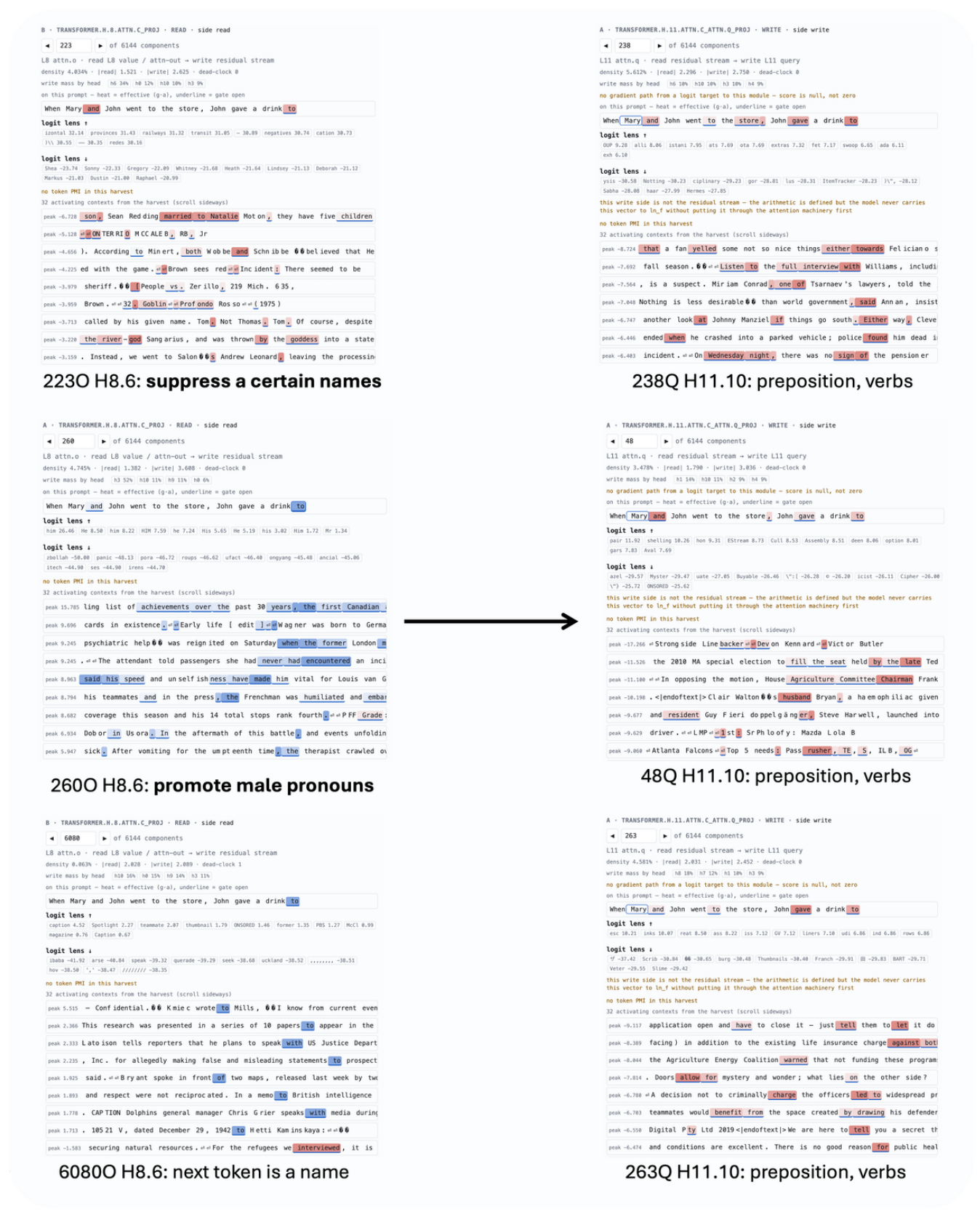}
    \caption{Connection between $H8.6$ (``S-inhibition head") and $H11.10$ (``negative name mover head"). Components $223O\, H8.6$ (``suppress certain names"), $260O\, H8.6$ (``promote male pronouns"), $6080O \, H8.6$ (``next token is a name"), which are the components that write the main functionality of $H8.6$ to the activation space, contribute strongly to the components $238Q, 48Q, 263Q \, H11.10$ (``prepositions, verbs").}
    \label{fig:v_in_h11_10}
\end{figure}

\begin{figure}[h]
    \centering
    \includegraphics[width=\linewidth]{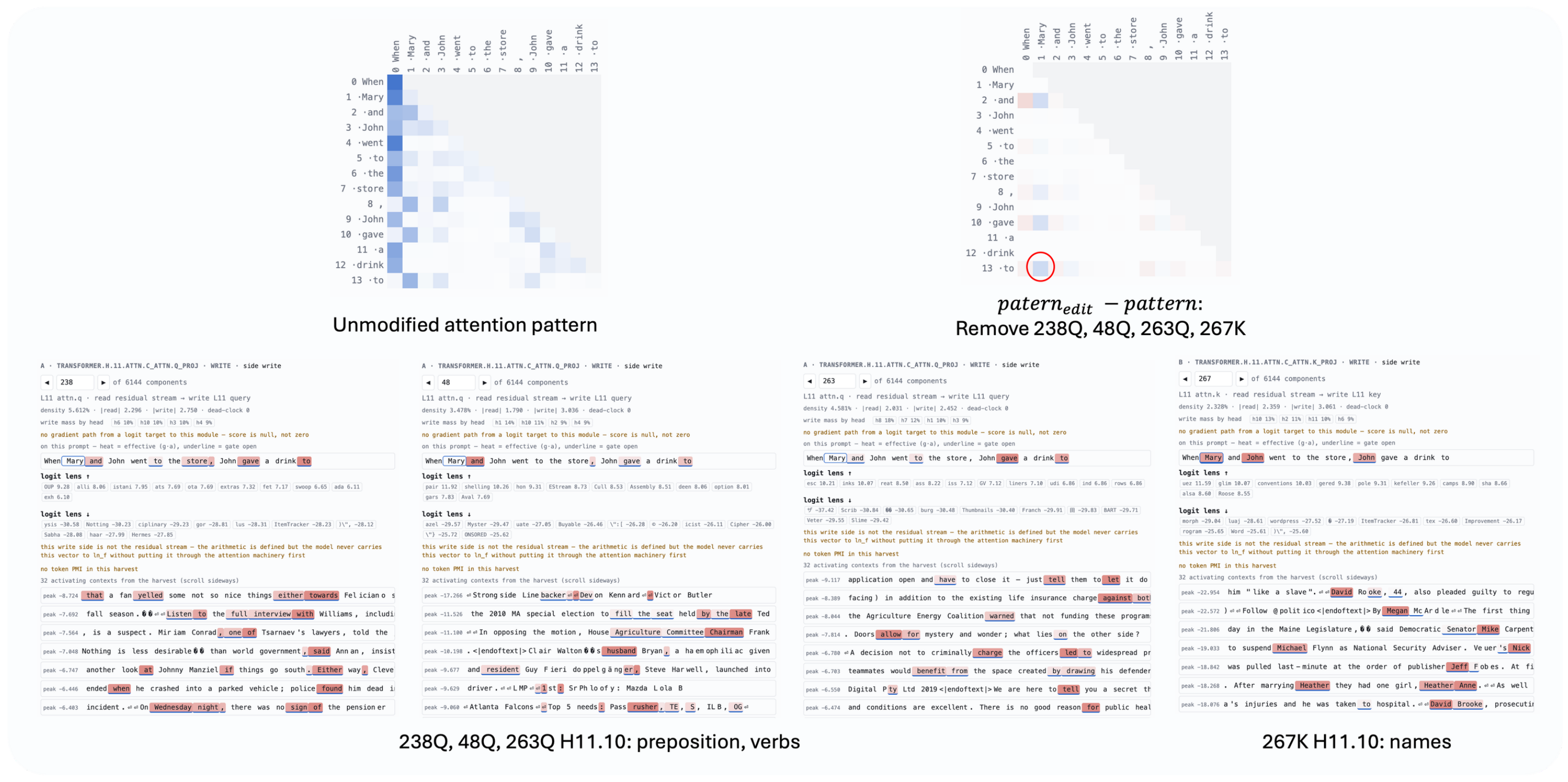}
    \caption{$QK$ circuit at position $Q:13(to)-K:1(Mary)$ of $H11.10$ (``negative name mover head"). We observed that components $238Q, 48Q, 263Q$ (``preposition, verbs"), $267K$ (``names") contribute strongly to the attention pattern, and ablating those components suppresses the ``negative name mover" attention pattern.}
    \label{fig:qk_h11_10}
\end{figure}

\begin{figure}[h]
    \centering
    \includegraphics[width=0.8\linewidth]{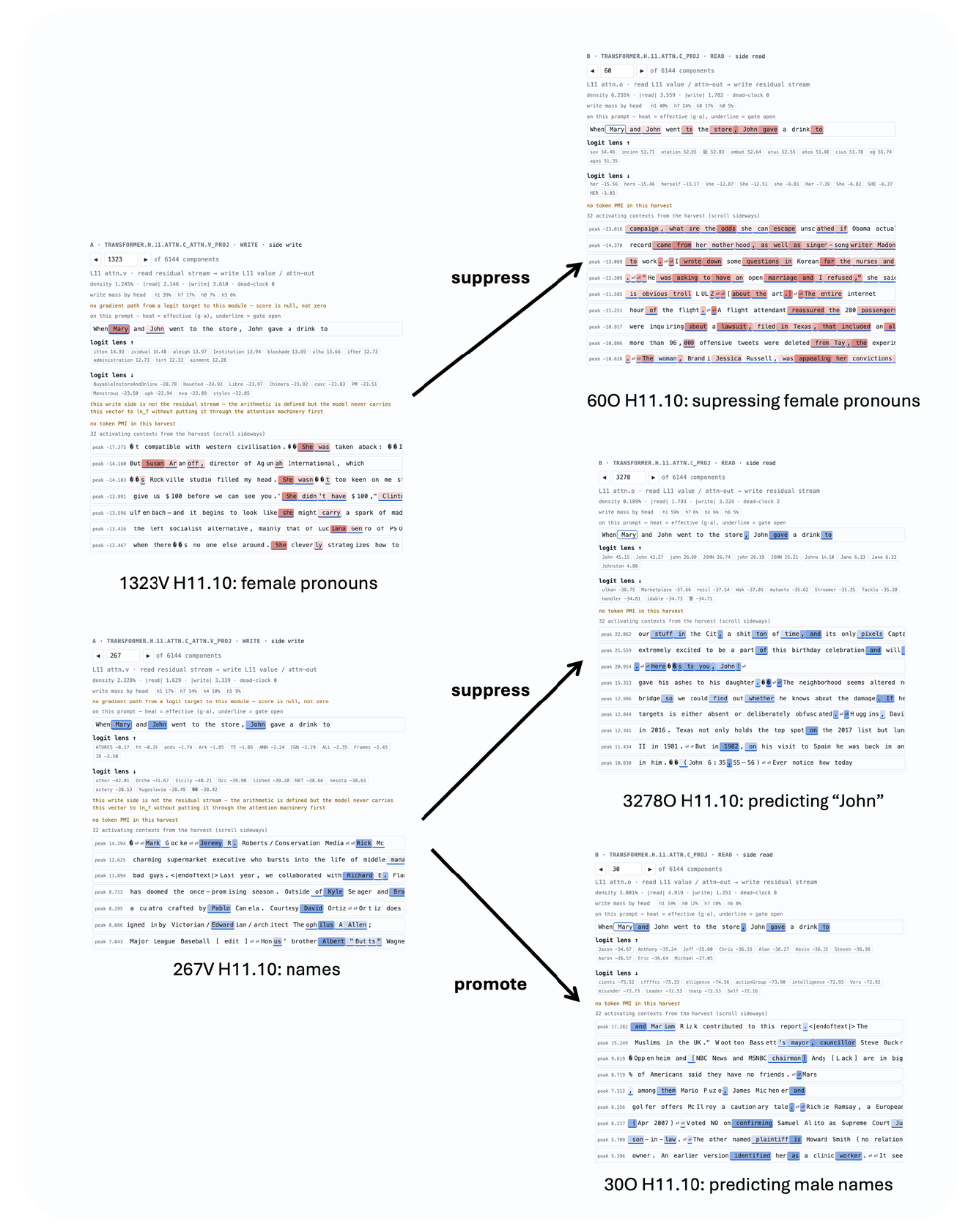}
    \caption{$OV$ circuit of $H11.10$ (``negative name mover head"). The components $1323V$ (``female pronouns"), $267V$ (``names") have a strong causal effect on $60O$ (``suppressing female pronouns"), $3278O$ (``predict John"), $30O$ (``predict male names"). The components $60O, 3278O, 30O$ promote/suppress ``male/female names/pronouns" and write the information to the activation space, which is the main role of $H11.10$.}
    \label{fig:ov_h11_10}
\end{figure}

\section{Behavioural probes: Finding ``Induction", ``Duplicate Token", ``Names and repeated names" Components}
\label{sec:behaviour-probes}

This section describes three probes that detects ``Induction", ``Duplicate Token", ``Names and
repeated names" components, supporting our argument in Appendix \ref{sec:ioi_circuit_full}.
Specifically, the three properties are \emph{induction}: a token is predictable because the sequence
repeats, \emph{duplication}: a token has occurred before, at no fixed offset, and \emph{names and
repeated names}: fires on all names but fires more on names that have occurred before.

\subsection{Computing probe score}
\label{sec:probe-design}

A component $c$ of a decomposed matrix $W$ contributes to that matrix's output through a single
scalar: the product of its gate and its activation,
\begin{equation}
  e_{t,c} \;=\; g_{t,c}\,a_{t,c}, \quad a_{t,c} = v_c^\top x_t
  \label{eq:effective}
\end{equation}
at token position $t$. Let $\mathcal{P}$ be the set of tokens that a component with the desired
behavior would fire on, and let $\mathcal{N}$ be the set of tokens that the component would not fire
if it contains the desired behavior, we compute the probe score:
\begin{equation}
  \mathrm{probe\_score}(c)
  \;=\;
  \frac{1}{\lvert \mathcal{P}\rvert}\sum_{t\in\mathcal{P}}\lvert e_{t,c}\rvert
  \;-\;
  \frac{1}{\lvert \mathcal{N}\rvert}\sum_{t\in\mathcal{N}}\lvert e_{t,c}\rvert .
  \label{eq:gap}
\end{equation}
High score means the component exhibits the behavior that we are looking for. 

\paragraph{Per-head filtering.} We want to report head specific component, however, the components in $W_O$ is the
concatenation of the per-head value vectors (so it is shared across all heads) and head $h$ owns the coordinate block
$\mathcal{B}_h = [\,d_h h,\; d_h(h{+}1)-1\,]$ where $d_h$ is the head
dimension. Thus we measure the share of a component's read direction lying in head $h$:
\begin{equation}
  mass_{h,c} \;=\;
  \bigl\lVert v_c[\mathcal{B}_h] \bigr\rVert_2^2 \big/ \bigl\lVert v_c \bigr\rVert_2^2 ,
  \qquad \textstyle\sum_h mass_{h,c} = 1 .
  \label{eq:headmass}
\end{equation}
When a behaviour is attributed to a particular head, we rank only the components with
$mass_{h,c} \ge 0.25$, i.e.\ those reading at least a quarter of their direction from that head. We report both rankings: unfiltered and filtered.

\subsection{Probe 1: induction}
\label{sec:probe-induction}

\paragraph{Prompt.} A sequence of $n$ uniformly random token ids, drawn from the interior of
the vocabulary, is concatenated with itself behind a BOS token,
\[
  \texttt{[eos]}\; r_1 \ldots r_n\; r_1 \ldots r_n ,
\]
where $r_i$ is a random token. At position $t$ of the second copy the token equals the one at $t-n$,
so an induction head attends from $t$ to $t-n+1$ - the token that followed last time. While an
induction component will activate on the second occurrence of the token only. $\mathcal{P}$ is the
set of the second copy of the token and $\mathcal{N}$ is the first copy. We run $n = 60$ over $4$
prompts, giving $\lvert\mathcal{P}\rvert = \lvert\mathcal{N}\rvert = 240$ tokens.

\subsection{Probe 2: duplicate token}
\label{sec:probe-duplicate}

\textbf{Prompt A: Duplicated tokens.} We construct the prompt as follow:
\[
  \texttt{[eos]}\; r_1 \ldots r_n\; r_{\pi(1)} \ldots r_{\pi(n)},
\]
where $r_i$ is a random token and $\pi$ a uniformly random permutation. The first half is a random
sequence of tokens while the second half is a random permutation of the first half. $\mathcal{P}$ is the second half sequence, and
$\mathcal{N}$ the first sequence. We run $n = 60$ over $4$ prompts, giving
$\lvert\mathcal{P}\rvert = \lvert\mathcal{N}\rvert = 240$ tokens.

\textbf{Prompt B: Duplicated names.} We fix a pool of $k = 12$ single-token first names, then append
$2n$ further names drawn uniformly from the same pool:
\[
  \texttt{[eos]}\; N_1 \ldots N_k\; \tilde N_1 \ldots \tilde N_{2n} ,
  \qquad \tilde N_j \sim \mathrm{Unif}\{N_1,\ldots,N_k\} .
\]
where $N_i$ is a name. $\mathcal{P}$ is the $2n$ resampled slots, and $\mathcal{N}$ is the $k$ pool slots. We run $n = 60$ over $4$ prompts.


\subsection{Probe 3: names and repeated names}
\label{sec:probe-factorial}

We test our hypothesis of a component fire on names and fire more strongly on the second occurrence of the name. 

\paragraph{Prompt A: random co-occurrence.} The prompt contains an interleave of a first name and common nouns, the names and nouns are drawn from two pools of $12$ items each:
\[
  \texttt{[eos]}\;\ \texttt{John}\ \ \texttt{table}\ \ \texttt{Mary}\ \ \texttt{river}\ \
  \texttt{John}\ \ \texttt{garden}\ \ldots
\]
Each slot is labelled by the token type (name or word) and by whether it is the first token or repeated token, giving the four classes $\texttt{name\_first}$, $\texttt{name\_rep}$,
$\texttt{word\_first}$, $\texttt{word\_rep}$. We test with $24$ names and nouns, on $4$ different prompts.

\paragraph{Prompt B: natural co-occurrence.} We repeat the same kind of prompt but using 6 hand-written passages to ensure a natural language token distribution, each containing one
first name and one common noun that recur naturally, e.g.
\begin{quote}\itshape
  The morning meeting ran long, and \textbf{John} brought \textbf{coffee} for everyone waiting
  there. By the time \textbf{John} sat down, the \textbf{coffee} had gone cold.
\end{quote}

\paragraph{Metric.} Let $\bar e_{\kappa}$ be the mean of $\lvert e_{t,c}\rvert$ over the
tokens of class $\kappa$, we measure the value:
\begin{align}
  \Pi_{\text{name}} &= \bar e_{\texttt{name\_rep}} - \bar e_{\texttt{name\_first}}, &
  \Pi_{\text{word}} &= \bar e_{\texttt{word\_rep}} - \bar e_{\texttt{word\_first}}, \notag \\[2pt]
  \Delta &= \Pi_{\text{name}} - \Pi_{\text{word}}. &&
  \label{eq:interaction}
\end{align}
This value will have: low value for a pure name detector component, low value for a pure duplicate token component; and $\Delta > 0$ for a duplicate-name detector where the second token is fired more strongly.

\section{Interaction of components, features}
\label{sec:coactivation}

\begin{figure}[h]
    \centering
    \includegraphics[width=1\linewidth]{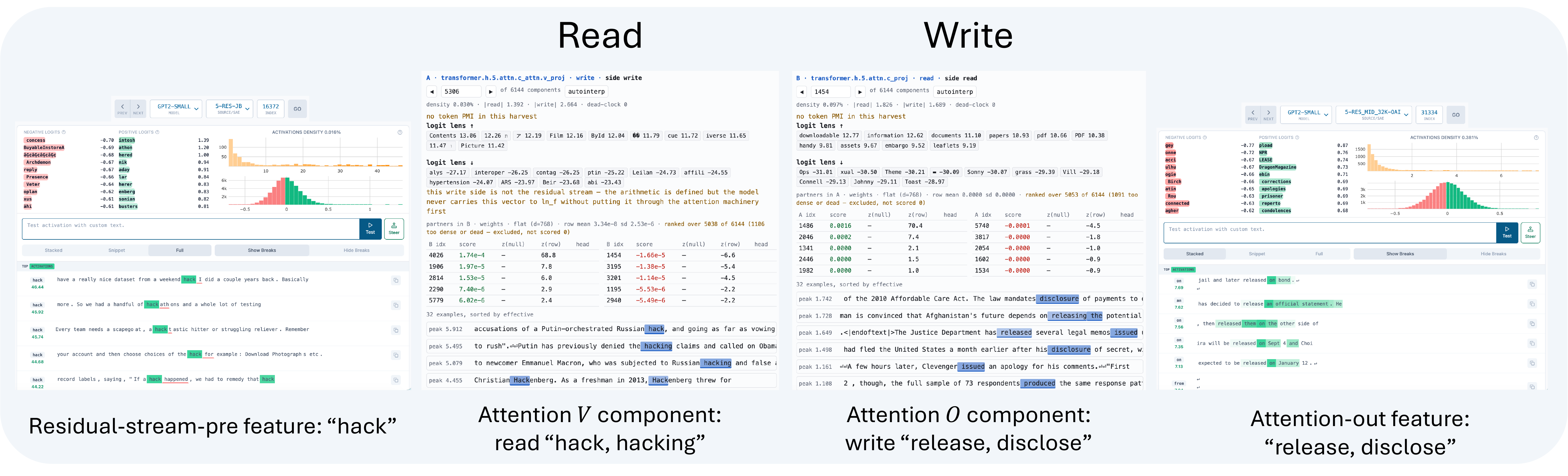}
    \caption{The image shows the component $5342 \, Q$ (``tech, car, airplane companies") at layer 9 reads the features $1964, 13889, 19129$ of residual-stream-pre which activate on ``tech, car companies" context. $5342 \, Q$ interacts with components $1518 \, K$ that fires on ``device and tech-related" and read the related features $18090, 2403, 2823$ from the residual-stream-pre, forming attention pattern.}
    \label{fig:knowledge_tech}
\end{figure}

\begin{figure}[h]
    \centering
    \includegraphics[width=1\linewidth]{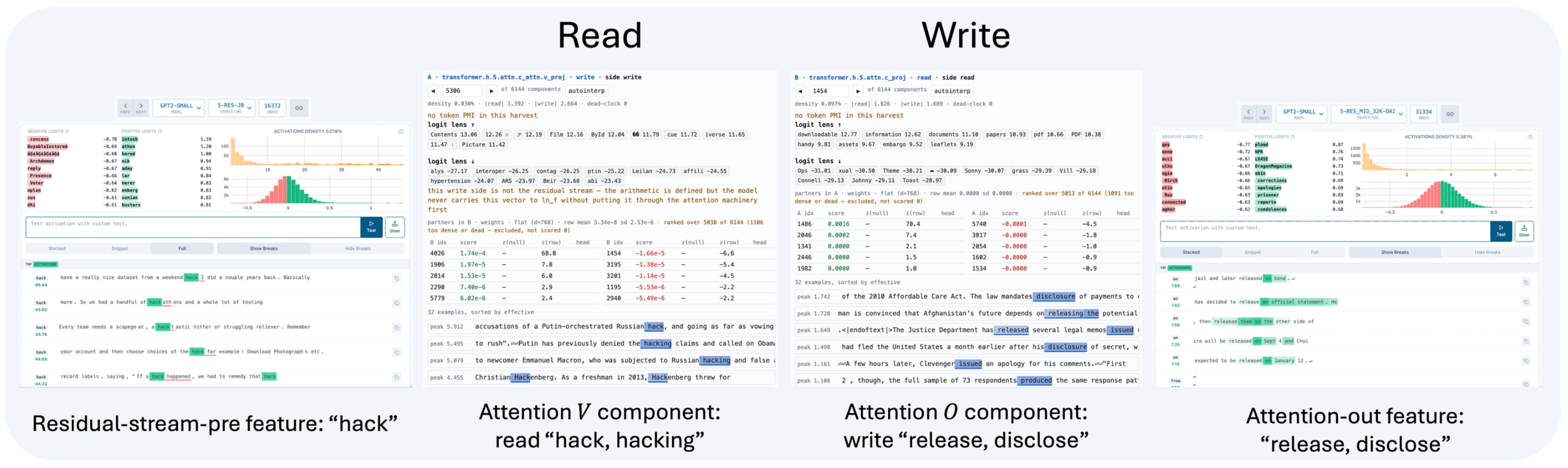}
    \caption{The image shows the component $5306 \, V$ (``hack, hacking") at layer 5 reads the feature $16372$ of residual-stream-pre which activates on ``hack" context. $5306 \, V$ interacts with components $1454 \, O$ that fires on ``release, disclose" and write out news related features $31334$ at the output of the attention module.}
    \label{fig:knowledge_hack}
\end{figure}

In this section, we outline how we compute the interaction between components and components, and between components and features. We also provide some additional interesting examples we saw in our implementation (Figure \ref{fig:knowledge_tech}, \ref{fig:knowledge_hack}).

\textbf{Component and component interaction.} The causal effect between two components $c_1, c_2$ is controlled by both the activation space at which they fire and their geometry, $u_{c_1}, v_{c_1}, u_{c_2}, v_{c_2}$. We therefore score a component pair by the corpus average of its first-order contribution rather than by pure cosine similarity.

Let $e_{t,c} \;=\; g_{t,c}\,a_{t,c}, \; a_{t,c} = v_c^\top x_t$. For interaction between $OV$ or $MLP_{in}, MLP_{out}$, at token $t$, component $c_1$ writes $e_{c_1}(x_t)\,u_{c_1}$ to the output; $c_2$ reads
$e_{c_1}(x_t)\,\langle u_{c_1},\,v_{c_2}\rangle$ the tokens at which ${c_2}$ fire.
The contribution of $c_1$ to ${c_2}$ at that token is therefore
$g_{t,{c_2}}\,e_{c_1}(x_t)\,\langle u_{c_1},\,v_{{c_2}}\rangle$:
\begin{equation}
  interact^{OV \, \text{or} \, MLP}({c_1},{c_2}) \;=\; \mathbb{E}_t\!\left[\, g_{t,{c_2}}\,e_{c_1}(x_t) \,\right]\;\big\langle u_{c_1},\,v_{{c_2}}\big\rangle .
  \label{eq:interaction_ov_mlp}
\end{equation}

For $QK$ circuit, both components write to the $QK$ space, let $c_q, c_k$ be the components in the $Q,K$ matrices respectively. The interaction is also divided by heads, therefore the interaction is:
\begin{equation}
  interact^{QK}_{h}(c_q,c_k) \;=\; \mathbb{E}_t\!\left[\, g_{t,{c_k}}\,e_{c_q}(x_t) \,\right]\;\cdot\; \big\langle u^{h}_{c_q},\,u^{h}_{c_k}\big\rangle ,
  \label{eq:interaction_qk}
\end{equation}
for $c_q$ and $c_k$ are component of $Q$ and $K$ respectively.

\textbf{Component and feature interaction.} For component and feature interaction, we measure the inner product between the directions: on the write side $ \langle u_c, W^{dec}_i \rangle$ for a feature downstream of the matrix, and on the read side $\langle v_c, W^{enc}_i\rangle$ for a feature upstream of it (with the intervening LayerNorm gain folded into $v_c$): 
\begin{equation}
    interact^{feature}_{downstream}(c,f_i) = \langle u_c, W^{dec}_i \rangle,  \qquad interact^{feature}_{upstream}(c,f_i) = \langle v_c, W^{enc}_i\rangle, 
\end{equation}

\section{Discussion: why we train ASPD's $g^s$ on the residual stream activation?}
\label{sec:discussion_gs}

In the training of ASPD (Appendix \ref{sec:training_details}), we train our shared causal importance function $g^s$ on the residual stream activation of the model. The reason for this is that training the function on the input activation $x$ of a weight can lead to a less diverse set of components and less interpretable components (an example of this is PD Transcoder on Gemma-2-2B, Table \ref{tab:interp}), because most components are ``near-dead'' components that activate on very few tokens that are unrelated. This happens when the input activation is inside the module, such as between the $MLP_{in}$ and $MLP_{out}$ or between the $O$ and $V$ matrices. We hypothesize that this phenomenon happens because the modules only process a small amount of information while discarding other information that is in the residual stream; hence, feeding $g^s$ with input activation with less information makes the decomposition process learn less diverse components, and most components are near-dead. Our current solution is to train the causal importance function entirely on the residual stream activation, feeding it with diverse information; we assume that if any activation space feature (or information) is not used by the weight, then the component that aligns with the feature will have a low $P_c$ norm. In practice, we found that this solution solves the problem by learning a diverse and interpretable component set while performing relatively the same on other metrics.

\end{document}